%% file: paper.tex
\documentclass[]{oppo}
\usepackage[toc,page,header]{appendix}
\input{common}

\usepackage{cleveref}
\theoremstyle{plain}

\theoremstyle{definition}

\theoremstyle{remark}

\usepackage[textsize=tiny]{todonotes}
\usepackage{fontawesome}

\usepackage{makecell}
\usepackage{xspace}
\usepackage{natbib}

\usepackage{wrapfig}

\newcommand{\obsbox}[1]{%
    \begin{tcolorbox}[colframe=black!70, colback=yellow!5, boxrule=1pt, arc=2mm,   top=2pt, bottom=3pt, left=3pt, right=3pt,
  boxsep=1pt]
        \small#1
    \end{tcolorbox}
}
\usepackage[table,xcdraw,usenames,dvipsnames]{xcolor}

\definecolor{codegreen}{rgb}{0,0.6,0}
\definecolor{codegray}{rgb}{0.5,0.5,0.5}
\definecolor{codepurple}{rgb}{0.58,0,0.82}
\definecolor{backcolour}{rgb}{0.95,0.95,0.92}

\lstdefinestyle{mystyle}{
    backgroundcolor=\color{backcolour},   
    commentstyle=\color{codegreen},
    keywordstyle=\color{magenta},
    numberstyle=\tiny\color{codegray},
    stringstyle=\color{codepurple},
    basicstyle=\ttfamily\footnotesize,
    breakatwhitespace=false,         
    breaklines=true,                 
    captionpos=b,                    
    keepspaces=true,                 
    numbers=left,                    
    numbersep=5pt,                  
    showspaces=false,                
    showstringspaces=false,
    showtabs=false,                  
    tabsize=2
}

\usepackage{twemojis}
\usepackage{fontawesome}
\usepackage{libertine}
\renewcommand{\ttfamily}{\fontfamily{cmtt}\selectfont}

\newcommand{\ourmethod}{{\fontfamily{lmtt}\selectfont \textbf{MemEvolve}}\xspace}

\newcommand{\ourframework}{{\fontfamily{lmtt}\selectfont \textbf{EvolveLab}}\xspace}

\title{MemEvolve: Meta-Evolution of Agent Memory Systems}

\affiliation{OPPO AI Agent Team, LV-NUS lab}

\abstract{
Self-evolving memory systems are unprecedentedly reshaping the evolutionary paradigm of large language model (LLM)-based agents. Prior work has predominantly relied on manually engineered memory architectures to store trajectories, distill experience, and synthesize reusable tools, enabling agents to evolve on the fly within environment interactions. However, this paradigm is fundamentally constrained by the \textit{staticity} of the memory system itself: while memory facilitates agent-level evolving, the underlying memory architecture cannot be meta-adapted to diverse task contexts.
To address this gap, we propose \ourmethod, a meta-evolutionary framework that jointly evolves agents’ experiential knowledge and their memory architecture, allowing agent systems not only to accumulate experience but also to progressively refine how they learn from it. To ground \ourmethod in prior research and foster openness in future self-evolving systems, we introduce \ourframework, a unified self-evolving memory codebase that distills twelve representative memory systems into a modular design space (\textit{encode}, \textit{store}, \textit{retrieve}, \textit{manage}), providing both a standardized implementation substrate and a fair experimental arena.
Extensive evaluations on four challenging agentic benchmarks demonstrate that \ourmethod achieves (I) substantial performance gains, improving frameworks such as SmolAgent and Flash-Searcher by up to $17.06\%$; and (II) strong cross-task and cross-LLM generalization, designing memory architectures that transfer effectively across diverse benchmarks and backbone models.

}

\date{\today}
\checkdata[\faGithub\;Code]{\url{https://github.com/bingreeky/MemEvolve}}
\begin{document}
\maketitle

\begin{figure}
    \centering
    \vspace{-2em}
    \includegraphics[width=\linewidth]{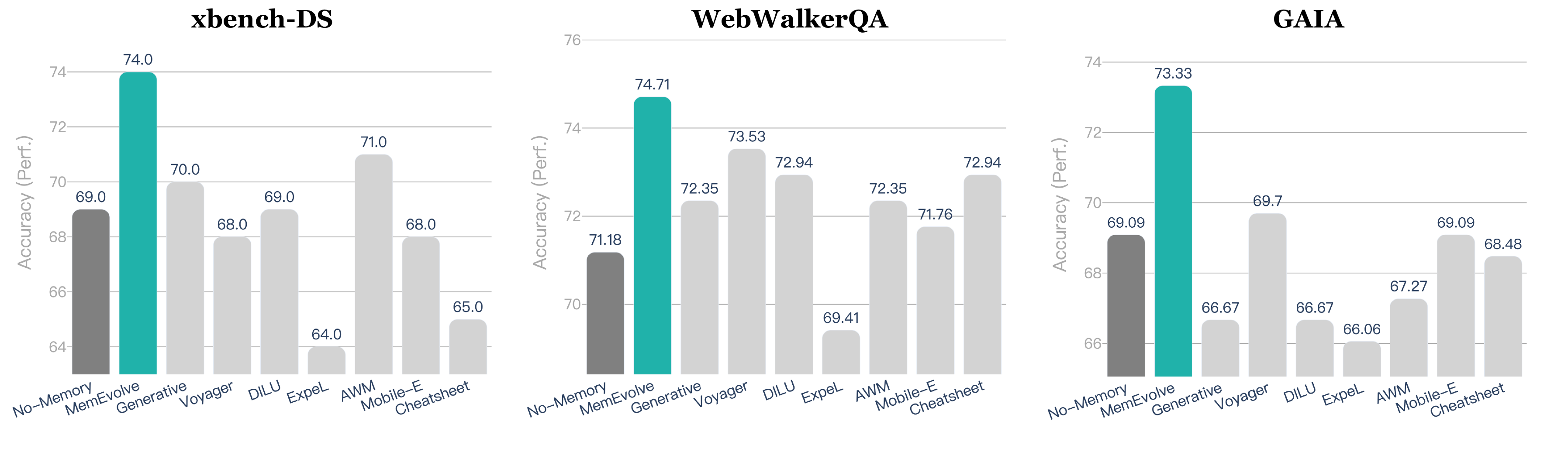}
    \vspace{-2em}
    \caption{The comparison between \ourmethod and several popular self-evolving agent memory systems across benchmarks. The underlying framework is Flash-Searcher~\citep{qin2025flashsearcherfasteffectiveweb}+\textsc{GPT-5-Mini}.}
    \label{fig:intro-1}
\end{figure}

\section{Introduction}
\label{sec:intro}

Language agents and agent systems, empowered by increasingly capable foundation models~\citep{kimiteam2025kimik2openagentic,meituanlongcatteam2025longcatflashtechnicalreport} and sophisticated scaffolding~\citep{openhands,lang2023langchain}, have advanced rapidly, demonstrating unprecedented performance across complex tasks such as deep research~\citep{chen2025xbenchtrackingagentsproductivity}, scientific discovery~\citep{bai2025interns1scientificmultimodalfoundation,wei2025aiscienceagenticscience}, and industrial report generation~\citep{zhang2025deepresearchsurveyautonomous}. A key driving force behind this success is the \textit{agent memory system}~\citep{RUC2024agent-memory,hu2025memoryageaiagentssurvey}, which persistently captures interactions between the agent and environment, distilling them into diverse forms of knowledge and skills, and thereby enabling large language model (LLM)-based agents to evolve continuously in task solving and world exploration~\citep{wu2025humanmemoryaimemory}.

Naturally, the choice of memory paradigm plays a decisive role in shaping an agent’s capacity for on-the-fly self-evolution. Initial designs centered on raw trajectory storage and few-shot prompting~\citep{zhong2024memorybank,wen2024diluknowledgedrivenapproachautonomous}, which were later superseded by more abstracted textual artifacts such as tips, shortcuts, and reasoning templates~\citep{ouyang2025reasoningbankscalingagentselfevolving,zhang2025gmemorytracinghierarchicalmemory,ye2025h2r,tang2025chemagentselfupdatinglibrarylarge}. Recent advances have also explored structured tool interfaces (\textit{e.g.}, APIs~\citep{zheng2025skillweaverwebagentsselfimprove}, MCPs~\citep{qiu2025alita,qiu2025agentdistilltrainingfreeagentdistillation,zhang2025agentorchestraorchestratinghierarchicalmultiagent}) and code-level repositories~\citep{zhang2025darwingodelmachineopenended,wang2025huxleygodelmachinehumanlevelcoding} as memory carriers. Amid this growing diversity, an inquisitive practitioner might ask: \textit{What kind of memory architecture most effectively drives agent self-improving?}

\begin{figure}
    \centering
    \includegraphics[width=\linewidth]{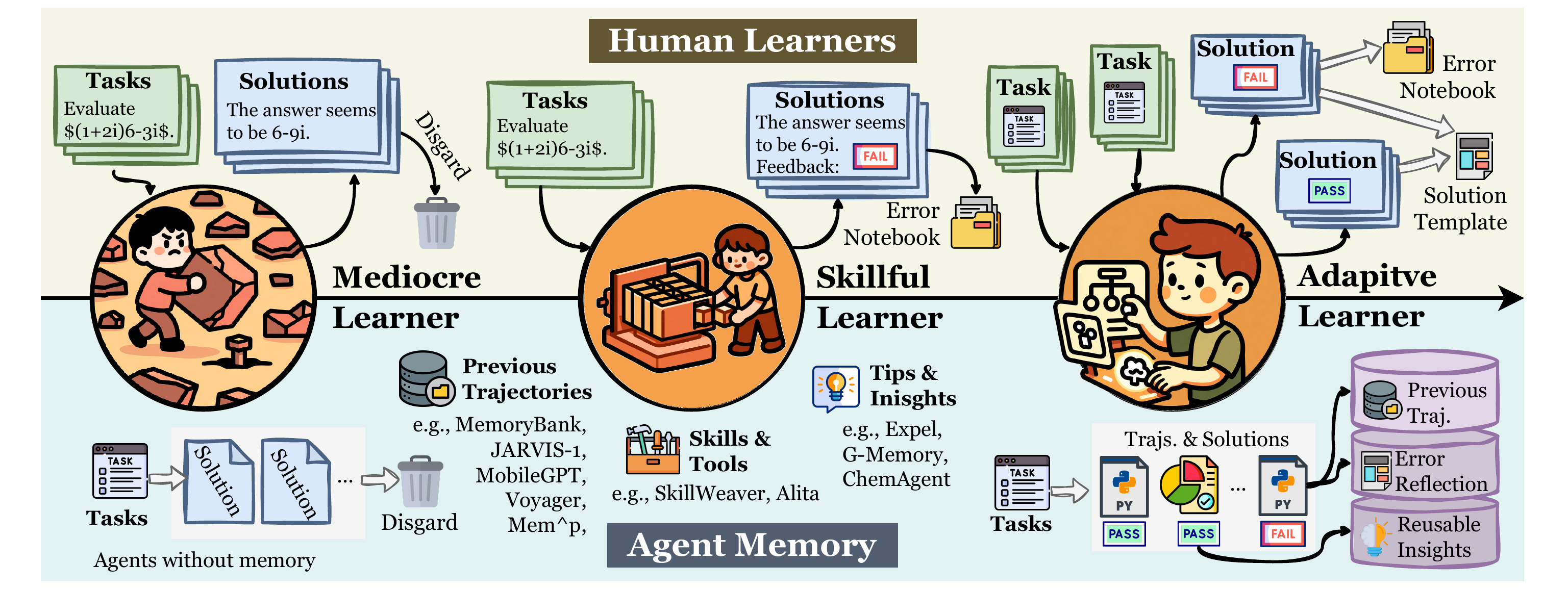}
    \vspace{-1.2em}
    \caption{The paradigm of agent self-evolution admits a natural analogy to human learning. At one extreme, a \emph{mediocre learner} fails to benefit from experience (agents without memory). More capable \emph{skillful learners} can extract reusable skills from past experience, albeit through a fixed and pre-defined abstraction scheme. In contrast, an \emph{adaptive learner} simultaneously accumulates experience and dynamically adjusts the strategy by which experience is consolidated and utilized. This final regime precisely characterizes the objective of \ourmethod.
}
    \label{fig:intro}
    \vspace{-0.9em}
\end{figure}

We posit that no universally optimal memory architecture exists. For instance, a memory system that distill reusable APIs from past trajectories may excel in tasks such as web browsing, yet offer limited utility for mathematical and scientific reasoning. Conversely, memories predicated on self-critique, while powerful in reasoning-intensive domains~\citep{cai2025trainingfreegrouprelativepolicy}, show diminished efficacy in coding and tool-use scenarios, as empirically discussed in \citep{zhang2025agentracerinducingfailurellm}. We contend that these trade-offs arise from the \textbf{static nature} of current memory systems. Researchers typically design a fixed memory pipeline (\textit{i.e.}, memory ingestion/abstraction/retrieval~\citep{zhang2025memengineunifiedmodularlibrary}) and embed it within an agent, assuming it will sustain long-term evolution through mere exposure to new experiences. Yet this overlooks a crucial reality: distinct tasks are coupled with distinct memory affordances. A memory system that cannot adapt itself to the task at hand is fundamentally misaligned with the very premise of open-ended agent evolution.

To elucidate this dilemma, consider the analogy of human learning. Both high- and low-performing students inevitably make mistakes, yet their distinction lies in the meta-cognitive strategies they employ to \textit{learn} from these errors. An underperforming student might resort to rote memorization, superficially recording an error without genuine comprehension~\citep{zhong2024memorybank,orhan2023recognitionrecallretentionfewshot}. In contrast, a more skillful student engages in higher-order learning: they not only record errors but also distill transferable insights through reflection~\citep{reflexion,zhao2024expel} or derive reusable schemas~\citep{zheng2025skillweaverwebagentsselfimprove,qiu2025alita}). Current memory systems effectively model a \textit{skillful} learner. Herein lies the critical gap: the most effective human learners are not merely skillful, but \textit{adaptive}. They dynamically alter their learning strategies based on the subjects, for instance, prioritizing memorization for literary analysis while abstracting solution templates for mathematics. It is precisely this transition, from a \textit{skillful} to an \textit{adaptive} learner (as shown in \Cref{fig:intro}), that we argue agent memory systems must undergo. To put it more formally:

\obsbox{How can a memory system not only facilitate the agent system's evolution but also \textit{meta-evolve its own architecture} to achieve superior task-domain performance gains while preserving generalizability?}

To address the challenge, we introduce \ourmethod, a framework that facilitates the dual evolution of an agent's experience and its memory architecture. Conceptually, \ourmethod operates as a bilevel optimization process: the inner loop performs a \textit{first-order evolution}, where the agent, guided by a fixed memory system, adapts to a continuous stream of new tasks by populating its experience base. The outer loop drives a \textit{second-order evolution}, meta-learning a more effective memory architecture to accelerate future learning. This allows the agent not only to evolve, but to evolve more efficiently and intelligently over time.

However, the vast and heterogeneous design space of memory systems (\textit{e.g.}, knowledge graphs, skill libraries, vector databases) presents a significant challenge to controllable optimization. To render this optimization tractable, we introduce a modular design, decomposing any memory architecture into four key components: \ding{168} \textit{Encode} (perceiving and formatting experiences), \ding{169} \textit{Store} (committing information), \ding{170} \textit{Retrieve} (context-aware recall), and \ding{171} \textit{Manage} (consolidation and forgetting).  \ourmethod evolves the programmatic implementations of these modules in a model-driven fashion, using feedback from the agent's performance in the inner loop. This process establishes a virtuous cycle: an improved memory architecture from the outer loop enhances the agent's learning efficiency. In turn, a more capable agent generates higher-quality trajectories, providing a more precise fitness signal for the outer loop to drive the next round of architectural evolution.

To ground our framework within the diverse landscape of existing self-improving agent memories, we systematically re-implement twelve representative architectures in a unified modular design space, including \textsc{ExpeL}~\citep{zhao2024expel}, \textsc{Agent Workflow Memory}~\citep{wang2024agentworkflowmemory}, and \textsc{Dynamic Cheatsheet}~\citep{suzgun2025dynamiccheatsheettesttimelearning}. The resulting framework, denoted as \ourframework, serves both as an empirical foundation for \ourmethod's evolutionary process and as a standardized codebase to facilitate future research on self-evolving agents. Our contributions are as follows:

\begin{itemize}[leftmargin=1.2em,itemsep=-0.2em]
\item[\ding{182}] \textbf{Unified Codebase:} We introduce \ourframework, a modular design space for self-improving agent memory systems encompassing four key components (\textit{encoding}, \textit{storage}, \textit{retrieval}, and \textit{management}), providing unified implementations and benchmark support for a wide range of prevailing agent memory systems.

\item[\ding{183}] \textbf{Meta-Evolution Framework:} We propose \ourmethod, a meta-evolutionary framework that jointly evolves both agents' experiential knowledge and their underlying memory architecture, in which agent systems not only accumulate experience but also progressively refine their mechanism for learning from it.

\item[\ding{184}] \textbf{Experimental Evaluation:} Extensive experiments on four challenging agentic benchmarks demonstrate that \ourmethod delivers (I) \textbf{substantial performance gains}, improving frameworks such as SmolAgent and Flash-Searcher by up to $17.06\%$; and (II) \textbf{cross-domain, cross-framework and cross-LLM generalization}, where memory systems evolved on TaskCraft yield $2.0-9.09\%$ gains with unseen benchmarks and backbone models.
\end{itemize}

\section{Related Work}
\label{sec:related}

\vspace{-0.4em}
\fakeparagraph{LLM Agent Systems} The past two years have witnessed rapid advances in LLM-based agent systems across multiple dimensions~\citep{tran2025multiagentcollaborationmechanismssurvey,fang2025comprehensivesurveyselfevolvingai}. In terms of \textbf{system complexity}, development has progressed from early single-agent setups with manually defined workflows and limited tool configurations~\citep{autogen,autogpt2023autogpt} to sophisticated multi-agent architectures featuring diverse MCP integrations and automated orchestration~\citep{zhang_aflow_2024,zhang2025evoflow,wang2025scoreflowmasteringllmagent,zhang2025maas}. From the perspective of \textbf{task domains}, capabilities have expanded from relatively constrained areas such as coding and mathematical reasoning~\citep{hong2024metagpt,yin2023exchangeofthoughtenhancinglargelanguage} to more challenging domains, including deep research and scientific discovery~\citep{du2025deepresearchbenchcomprehensivebenchmark,ghareeb2025robinmultiagentautomatingscientific}. Today, numerous open-source multi-agent systems demonstrate competitive performance on demanding benchmarks such as {GAIA}~\citep{mialon2023gaia}, HLE~\citep{phan2025hle}, {BrowseComp}~\citep{wei2025browsecomp}, and {xBench}~\citep{chen2025xbenchtrackingagentsproductivity}, including CAMEL's \textsc{OWL}~\citep{hu2025owloptimizedworkforcelearning}, Tencent's \textsc{CK-Pro}~\citep{fang2025cognitivekernelproframeworkdeep}, Skywork's \textsc{AgentOrchestra}~\citep{zhang2025agentorchestrahierarchicalmultiagentframework}, and ByteDance's \textsc{AIME}~\citep{shi2025aime}, among others.

\vspace{-0.4em}
\fakeparagraph{Agent Memory Architectures}  
Agent memory systems can be broadly divided by objective into \textit{personalized memory} and \textit{self-improving memory}~\citep{RUC2024agent-memory,hu2025memoryageaiagentssurvey}. The former enables agent chatbots to dynamically capture user-specific information and preferences, while the latter focuses on distilling knowledge and skills from continual interactions with the environment to enhance performance, a focus adopted in this work.
Self-improving memories are primarily differentiated by their storage modality. Early systems stored raw agent trajectories as few-shot examples~\citep{voyager,zhong2024memorybank,packer2023memgpt}; subsequent designs abstracted these experiences into higher-level lessons, insights~\citep{yang2025learningjobexperiencedrivenselfevolving,sun2025hierarchicalmemoryhighefficiencylongterm,wu2025evolverselfevolvingllmagents}, procedural tips~\citep{wang2025mobileagenteselfevolvingmobileassistant,zheng2025skillweaverwebagentsselfimprove,fang2025mempexploringagentprocedural}, and more recently, reusable tools and structured repositories~\citep{zhao2025pyvisionagenticvisiondynamic,qiu2025agentdistilltrainingfreeagentdistillation,qiu2025alita,zhang2025darwingodelmachineopenended}. Despite their differences in representation, there approaches share the same ambition, \textit{i.e.}, to enable agents to learn, adapt, and improve in a human-esque manner.

\section{EvolveLab: A Unified Codebase for Self-Evolving Memory}
\label{sec:evovelab}
\vspace{-0.4em}
In this section, we first formalize the LLM-based agentic system and its associated memory architecture, then present the modular design space of \ourframework, which comprehensively captures the characteristics of existing self-evolving agent memories, and finally introduce the unified codebase \ourframework.

\vspace{-0.4em}
\subsection{Preliminary}
\label{sec:preliminary}

We formalize an LLM-based agentic system as 
\(\mathcal{M} = \langle \mathcal{I}, \mathcal{S}, \mathcal{A}, \Psi, \Omega \rangle\), 
where \(\mathcal{I}\) indexes the $\{1,\cdots, N\}$ agents, \(\mathcal{S}\) denotes the shared state space, 
\(\mathcal{A} = \bigcup_{i \in \mathcal{I}} \mathcal{A}_i\) represents the joint action space, 
and \(\Psi(s_{t+1} \mid s_t, a_t, \mu(t))\) describes the environment dynamics with 
\(\mu(t) \in \mathcal{I}\) indicating the active agent at time step \(t\). 
The system leverages a memory module \(\Omega\), which maintains a continuously evolving memory state \(M_t\).
At each step, the active agent observes the current state \(s_t\), considers a task-specific query \(\mathcal{Q}\), 
and interacts with $\Omega$ to retrieve contextually relevant memory \(c_t\), 
conditioned on its interaction history \(\mathcal{H}_t\). The agent $\mu_t$'s policy $\pi_{\mu_t}$ then delivers an action:
\[
a_t = \pi_{\mu(t)}(s_t, \mathcal{H}_t, \mathcal{Q}, c_t),\;c_t \sim \Omega(M_t, s_t, \mathcal{H}_t, \mathcal{Q}).
\]

Following task execution, a trajectory \(\tau = (s_0, a_0, \dots, s_T)\) is recorded, 
with an overall performance evaluated via a terminal reward \(R(\tau)\). 
The memory system assimilates new experience units \(\epsilon\), which can vary in granularity 
(from individual state-action transitions to aggregated segments or complete trajectories), 
and updates the memory state as
\[
M_{t+1} = \Omega(M_t, \epsilon),
\]
where \(\Omega\) abstracts the memory’s mechanisms for integrating and organizing 
new experiences or knowledge. 


\subsection{Modular Design Space of Memory Systems}
\label{sec:modular}
The heterogeneous and rapidly evolving landscape of self-improving agent memories presents challenges for systematic analysis and controlled experimentation. To address this, we propose a modular design space that decomposes any memory system \(\Omega\) into four functionally distinct yet interdependent components: $
\Omega = (\mathcal{E}, \mathcal{U}, \mathcal{R}, \mathcal{G})$, representing \textit{encode}, \textit{store}, \textit{retrieve}, and \textit{manage} operations, respectively.

\begin{itemize}[leftmargin=1.5em,itemsep=-0.3em]
    \item \textbf{Encode (\( \mathcal{E} \))}: Transforms raw experiences, such as trajectory segments \(\tau_t = (s_t, a_t, s_{t+1})\), tool outputs, or self-critiques, into structured representations \(e_t = \mathcal{E}(\epsilon_t)\). Encoding may be as simple as compressing raw traces~\citep{zheng2023synapse} or as sophisticated as extracting generalizable lessons~\citep{zheng2025skillweaverwebagentsselfimprove}.
    \item \textbf{Store (\( \mathcal{U} \))}: Integrates encoded experiences into the persistent memory \(M_t\), yielding $M_{t+1} = \mathcal{U}(M_t, e_t)$.
    Storage can be vector databases~\citep{zhao2024expel}, knowledge graphs~\citep{zhang2025gmemorytracinghierarchicalmemory,rasmussen2025zeptemporalknowledgegraph}, or others.
    \item \textbf{Retrieve (\( \mathcal{R} \))}: Provides task-relevant memory content, formalized as $c_t = \mathcal{R}(M_t, s_t, \mathcal{Q})$,    which informs the agent’s policy decision \(a_t\). Retrieved content may include reusable tools~\citep{zhang2025agentorchestrahierarchicalmultiagentframework}, planning experience~\citep{tang2025chemagentselfupdatinglibrarylarge}, or distilled procedural knowledge~\citep{wu2025evolverselfevolvingllmagents,yang2025learningjobexperiencedrivenselfevolving,fang2025mempexploringagentprocedural}.
    \item \textbf{Manage (\( \mathcal{G} \))}: Performs offline and asynchronous operations such as consolidation, abstraction, or selective forgetting to maintain long-term memory quality and efficiency, denoted as $M'_t = \mathcal{G}(M_t)$.
\end{itemize}

This modular abstraction allows us to represent each memory system as a specific combination of programmatic implementations for \( (\mathcal{E}, \mathcal{U}, \mathcal{R}, \mathcal{G}) \), forming a ``genotype'' that facilitates the meta-evolutionary process of \ourmethod.

\begin{table*}[!t]
\centering
\footnotesize
\caption{A taxonomy of self-improving agent memory systems implemented in \ourframework. In the ``Mul.'' column, \faMale~indicates support for single-agent settings, while \faMale\faMale~denotes compatibility with multi-agent systems. ``Gran.'' specifies the granularity at which memory is provided (\textit{step-wise} vs.\ \textit{trajectory-wise}), and ``Online'' indicates whether memory is updated \textit{on-the-fly} ({\color{ForestGreen}\faChain}) or maintained as an offline experience repository ({\color{RedOrange}\faChainBroken}).
}
\vspace{-0.4em}
\label{tab:memory_systems_comparison}
\setlength{\tabcolsep}{3pt} 
\begin{tabular}{@{}l l|ccc|l l l l@{}}
\toprule
\textbf{Method} & \textbf{Date} & \textbf{Mul.} & \textbf{Gran.} & \textbf{Online} & \textbf{Encode} & \textbf{Store} & \textbf{Retrieve} & \textbf{Manage} \\
\midrule
I. {Voyager} & 2023.5 & \faMale & traj.  & {\color{ForestGreen}\faChain} & Traj. \& Tips & Vector DB & Semantic Search & N/A \\
II. {ExpeL} & 2023.8 & \faMale & traj.  & {\color{ForestGreen}\faChain} & Traj. \& Insights & Vector DB & Contrastive Comparison & N/A \\
III. {Generative} & 2023.10 & \faMale\faMale & traj.  & {\color{ForestGreen}\faChain} & Traj. \& Insights & Vector DB & Semantic Search & N/A \\
IV. {DILU} & 2024.2 & \faMale & traj.  & {\color{ForestGreen}\faChain} & Traj. & Vector DB & Semantic Search & N/A \\
V. {AWM} & 2024.9  & \faMale & traj. & {\color{ForestGreen}\faChain}{\color{RedOrange}\faChainBroken} & Workflows & Vector DB & Semantic Search & N/A \\
VI. {Mobile-E} & 2025.1 & \faMale & step  & {\color{RedOrange}\faChainBroken} & Tips \& Shortcuts & Vector DB & Semantic Search & N/A \\
VII. {Cheatsheet} & 2025.4 & \faMale & traj.  & {\color{ForestGreen}\faChain} & Tips \& Shortcuts & JSON & Semantic Search & N/A \\
VIII. {SkillWeaver} & 2025.4 & \faMale& traj.  & {\color{RedOrange}\faChainBroken} & APIs & Tool Library & Function Matching & Skill Pruning \\
IX. {G-Memory} & 2025.6 & \faMale\faMale & traj.  & {\color{ForestGreen}\faChain} & Tips \& Workflow & Graph & Graph/Semantic Search & Episodic Consolidation \\
X. {Agent-KB} & 2025.7 & \faMale\faMale & step & {\color{RedOrange}\faChainBroken} & Tips \& Workflow & Hybrid DB & Hybrid Search & Deduplication \\
XI. {Memp} & 2025.8 & \faMale & step.  & {\color{ForestGreen}\faChain} & Tips \& Workflow & JSON & Semantic Search & Failure-driven Adjustment \\
XII. {EvolveR} & 2025.10 & \faMale & step.  & {\color{ForestGreen}\faChain} & Tips \& Workflow & JSON & Contrastive Comparison & Update \& Pruning\\
\bottomrule
\end{tabular}
\vspace{-0.8em}
\end{table*}

\subsection{EvolveLab Codebase}
Based on the above design space, we introduce \ourframework, a unified and extensible codebase designed for the systematic implementation and evaluation of self-evolving memories, serving as a standardized resource for the community.

\vspace{-0.4em}
\fakeparagraph{Implementation}
The cornerstone of \ourframework is its modular and hierarchical design. Every memory architecture re-implemented in our codebase (see \Cref{tab:memory_systems_comparison}) inherits from a singular abstract base class, \textsc{BaseMemoryProvider}, which enforces the unified four-component interface: \ding{168} {Encode}, \ding{169} {Store}, \ding{170} {Retrieve}, and \ding{171} {Manage}. This ensures that diverse memory mechanisms can be managed, modified, and evolved under a consistent programmatic structure. More details on the implementations can be found at \Cref{app:EvolveLab}.

\vspace{-0.4em}
\fakeparagraph{Evaluation}
Beyond unified implementation, \ourframework provides a standardized testbed for rigorously assessing memory architectures across diverse agentic tasks. The framework offers out-of-the-box support for multiple challenging benchmarks, including {GAIA}~\citep{mialon2023gaia}, {xBench}~\citep{chen2025xbenchtrackingagentsproductivity}, and {DeepResearchBench}~\citep{du2025deepresearchbenchcomprehensivebenchmark}. \ourframework accommodates two evaluation paradigms: an \ding{110} \textbf{online} mode, where the experiential memory base is updated on-the-fly as the agent system processes a continuous stream of tasks, and an \ding{110} \textbf{offline} mode, where the memory system first accumulates experience from a static set of trajectories before being assessed on separate, unseen tasks. To ensure robust and versatile assessment, we support multiple evaluation protocols, including exact string matching and flexible LLM-as-a-Judge.

\section{MemEvolve: A Meta-Evolving Memory Framework}
\label{sec:method}

\subsection{Dual-Evolution Process}
\label{subsec:dual_evolution}

Traditional self-improving memory systems operate under a \textit{fixed memory architecture}, 
where the memory interface \(\Omega\) is predefined and remains static.
Within this architecture, the agent iteratively populates and updates its memory state \(M_t\) 
through interaction with the environment and task experiences. 
For a trajectory \(\tau\) induced by a query \(\mathcal{Q}\), the memory evolution follows
\[
M_{t+1} = \Omega(M_t, \epsilon_\tau), 
\quad 
\epsilon_\tau \in \mathcal{E}(\tau),
\]
where \(\mathcal{E}(\cdot)\) denotes an experience extraction operator that maps a trajectory to a set of experience units, 
and \(\epsilon_\tau\) is an element sampled from this set.
While this enables the accumulation of knowledge, it fundamentally precludes \emph{architectural adaptation}, as the memory interface \(\Omega\) itself remains immutable.

\begin{figure}
    \centering
    \includegraphics[width=\linewidth]{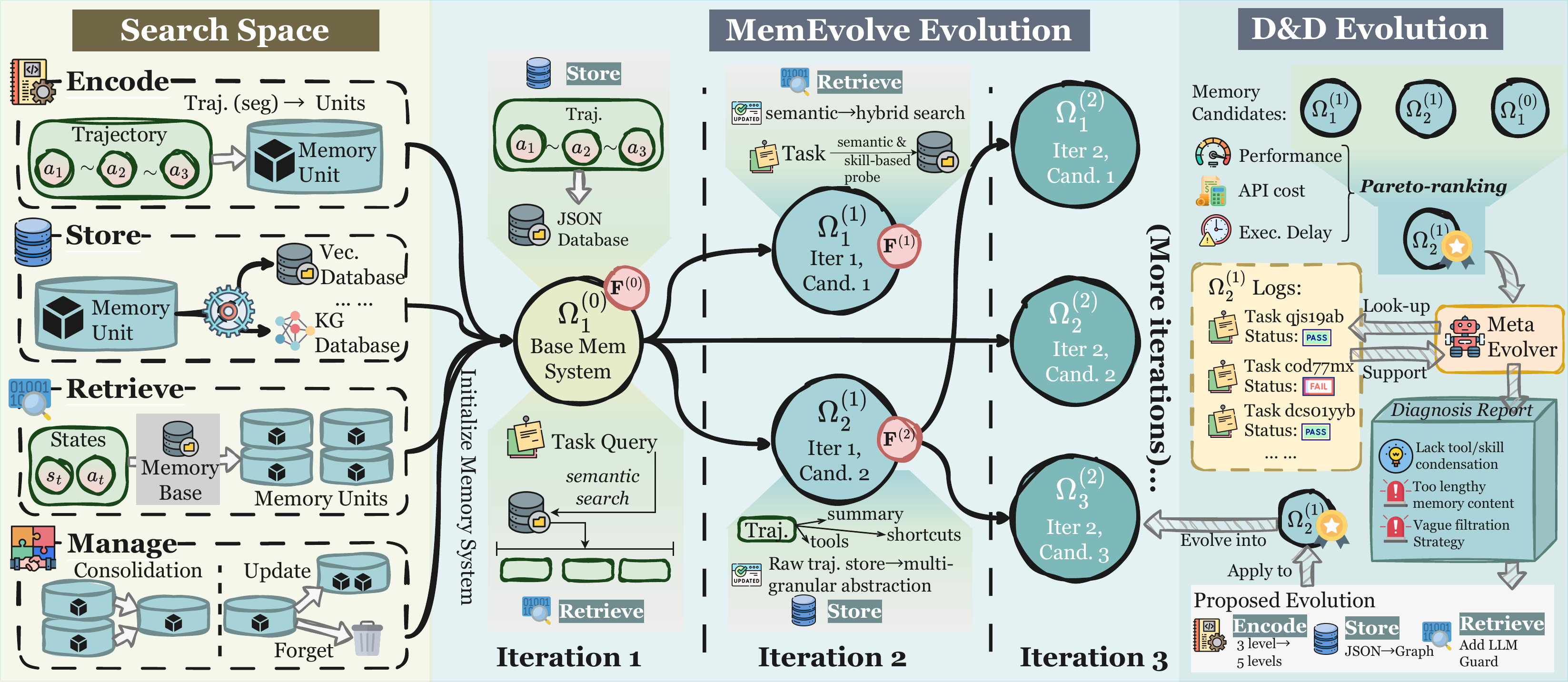}
    \caption{The overview of our proposed \ourmethod.}
    \label{fig:framework}
    \vspace{-1.4em}
\end{figure}

To transcend this limitation, we propose a \textbf{dual-evolution process} that jointly evolves 
(i) the agent’s memory base and (ii) the underlying memory architectures (as illustrated in \Cref{fig:framework}).
Instead of a single static \(\Omega\), we maintain, at each evolutionary iteration \(k\), 
a finite set of candidate memory systems \(\{\Omega_j^{(k)}\}_{j \in \mathcal{J}^{(k)}}\),
where each \(\Omega_j^{(k)}\) is instantiated as a concrete realization of the four-component memory interface $\Omega_j^{(k)} \triangleq \big(\mathcal{E}_j^{(k)}, \mathcal{U}_j^{(k)}, \mathcal{R}_j^{(k)}, \mathcal{G}_j^{(k)}\big)$.
The initial iteration start from a singleton set \(|\mathcal{J}^{(0)}| = 1\), corresponding to a hand-designed baseline memory, while later iterations admit multiple competing candidates. Given a batch of trajectories \(\mathcal{T}_j^{(k)}\) independently generated by executing the agent with memory system \(\Omega_j^{(k)}\), 
the dual-evolution process consists of two nested loops:

\begin{itemize}[leftmargin=1.5em,itemsep=-0.3em]
    \item \textbf{Inner Loop (Experience Evolution).} 
    For each candidate memory system \(\Omega_j^{(k)}\), the associated memory state \(M_{t,j}^{(k)}\), 
    initialized as an empty memory at the beginning of iteration \(k\), 
    is updated along trajectories \(\tau \in \mathcal{T}_j^{(k)}\) via
    \[
    M_{t+1,j}^{(k)} = \Omega_j^{(k)}\big(M_{t,j}^{(k)}, \epsilon_{\tau}\big),
    \quad 
    \epsilon_{\tau} \in \mathcal{E}_j^{(k)}(\tau).
    \]
    Executing the agent with \(\Omega_j^{(k)}\) over \(\mathcal{T}_j^{(k)}\) yields, for each trajectory \(\tau\), 
    a feedback vector \(\mathbf{f}_j(\tau) \in \mathbb{R}^d\), where \(d=3\) corresponds to the number of evaluation metrics 
    (\textit{i.e.}, task success, token consumption, and latency).
    An aggregation operator \(\mathcal{S}\) summarizes the inner-loop outcomes for each candidate as
    \[
    \mathbf{F}_j^{(k)} = \mathcal{S}\big(\{\mathbf{f}_j(\tau)\}_{\tau \in \mathcal{T}_j^{(k)}}\big),
    \quad
    j \in \mathcal{J}^{(k)}.
    \]
    
    \item \textbf{Outer Loop (Architectural Evolution).} 
    The set of memory architectures is then updated based on the collection of summaries 
    \(\{\mathbf{F}_j^{(k)}\}_{j \in \mathcal{J}^{(k)}}\).
    A meta-evolution operator \(\mathcal{F}\) selects high-performing candidates and proposes new variants, producing the next iteration's candidate set:
    \[
    \big\{\Omega_{j'}^{(k+1)}\big\}_{j' \in \mathcal{J}^{(k+1)}} 
    = 
    \mathcal{F}\!\left(
        \{\Omega_j^{(k)}\}_{j \in \mathcal{J}^{(k)}},\;
        \{\mathbf{F}_j^{(k)}\}_{j \in \mathcal{J}^{(k)}}
    \right).
    \]
    Specifically, \(\mathcal{F}\) ranks candidates according to \(\mathbf{F}_j^{(k)}\), 
    retains the top-\(K\) memory systems, and generates new architectures by modifying or recombining 
    all four components \((\mathcal{E}, \mathcal{U}, \mathcal{R}, \mathcal{G})\) of the selected candidates,
    where \(K\) denotes a fixed survivor budget.
    We detail the implementation of \(\mathcal{F}(\cdot)\) in \Cref{sec:dd-evolve}.
\end{itemize}

\noindent\textbf{Unified view.}
At a higher level, each iteration \(k\) alternates between 
(i) evolving the \emph{memory experience base} from an empty initialization under a fixed set of architectures, 
and (ii) evolving the \emph{memory architectures} themselves based on the induced performance:
\[
\big(\{\varnothing\}_{j \in \mathcal{J}^{(k)}},\, \{\Omega_j^{(k)}\}_{j \in \mathcal{J}^{(k)}}\big)
\;\xrightarrow{\text{inner}}\;
\big(\{M_{t+1,j}^{(k)}\}_{j \in \mathcal{J}^{(k)}},\, \{\Omega_j^{(k)}\}_{j \in \mathcal{J}^{(k)}}\big)
\;\xrightarrow{\text{outer}}\;
\big(\{M_{t+1,j}^{(k)}\}_{j \in \mathcal{J}^{(k)}},\, \{\Omega_{j'}^{(k+1)}\}_{j' \in \mathcal{J}^{(k+1)}}\big).
\]
By iterating this dual-evolution process, the agent does not merely accumulate experience within a fixed memory system; instead, both the memory base and the governing memory architectures co-evolve, yielding increasingly adaptive and resource-aware memory-driven behavior over time.

\subsection{Diagnose-and-Design Evolution}
\label{sec:dd-evolve}

We now detail the meta-evolution operator \(\mathcal{F}\), which governs the architectural update in each evolutionary iteration. Conceptually, \(\mathcal{F}\) decomposes into two coordinated components: (i) \emph{architectural selection}, which identifies a subset of high-performing memory systems to serve as evolutionary parents, and (ii) \emph{diagnose-and-design evolution}, which generates new memory architectures from each selected parent through a structured diagnosis procedure followed by a constrained redesign within the modular memory design space.

\fakeparagraph{Architectural Selection}
Given the candidate set \(\{\Omega_j^{(k)}\}_{j\in\mathcal{J}^{(k)}}\) and their corresponding summaries \(\{\mathbf{F}_j^{(k)}\}\), 
we define each summary vector as
\[
\mathbf{F}_j^{(k)} \triangleq 
\big(
\text{Perf}_j^{(k)},\;
-\text{Cost}_j^{(k)},\;
-\text{Delay}_j^{(k)}
\big),
\]
where higher values are preferred in all dimensions.
Candidates are first ranked by non-dominated sorting over \(\mathbf{F}_j^{(k)}\), yielding a Pareto rank \(\rho_j^{(k)}\).
Within the same Pareto rank, candidates are further ordered by the primary performance metric \(\text{Perf}_j^{(k)}\).
The top-\(K\) candidates are selected as the parent set:
\[
\mathcal{P}^{(k)} =
\operatorname*{Top\text{-}K}_{j \in \mathcal{J}^{(k)}}
\Big(\rho_j^{(k)},\; \text{Perf}_j^{(k)}\Big).
\]
This selection step ensures that architectural evolution is guided by systems that exhibit favorable trade-offs between task effectiveness and resource efficiency, while prioritizing task performance among Pareto-equivalent candidates.

\fakeparagraph{Diagnose-and-Design Evolution}
For each parent architecture \(\Omega_p^{(k)} \in \mathcal{P}^{(k)}\), \(\mathcal{F}\) generates a set of \(S\) descendants \(\{\Omega_{p,s}^{(k+1)}\}_{s=1}^S\) through a two-phase process:

\begin{itemize}[leftmargin=1.5em,itemsep=-0.3em]
    \item \textbf{Diagnosis.}
    Each parent architecture is examined using trajectory-level evidence from its own execution batch \(\mathcal{T}_p^{(k)}\).
    For each trajectory, the agent provides outcome statistics (e.g., success indicators, token costs) together with a structured description of the associated task query.
    A replay interface grants access to the corresponding trajectories \(\tau \in \mathcal{T}_p^{(k)}\), enabling targeted inspection of memory behavior, including retrieval failures, ineffective abstractions, or storage inefficiencies.
    The diagnosis phase thus produces a structured defect profile \(\mathcal{D}(\Omega_p^{(k)})\), characterizing architectural bottlenecks across the four memory components
    \(\big(\mathcal{E}_p^{(k)}, \mathcal{U}_p^{(k)}, \mathcal{R}_p^{(k)}, \mathcal{G}_p^{(k)}\big)\).
    
    \item \textbf{Design.}
    Conditioned on the defect profile \(\mathcal{D}(\Omega_p^{(k)})\), a redesigned architecture is constructed by modifying only the permissible implementation sites within the modular interface, thereby ensuring compatibility and isolating architectural changes to the designated design space.
    The design step produces \(S\) variants by instantiating distinct but valid configurations of the four components:
    \[
    \Omega_{p,s}^{(k+1)} 
    = 
    \operatorname{Design}\!\left(
    \Omega_p^{(k)},\,
    \mathcal{D}(\Omega_p^{(k)}),\,
    s
    \right),
    \quad s \in \{1,\dots,S\}.
    \]
    These variants differ in encoding strategies, storage rules, retrieval constraints, or management policies, yet all conform to the unified memory-system interface and remain executable by the agent.
\end{itemize}

\noindent\textbf{Resulting update.}
Aggregating all descendants across parents yields the next set of candidate architectures:
\[
\{\Omega_{j'}^{(k+1)}\}_{j' \in \mathcal{J}^{(k+1)}} 
= 
\bigcup_{\Omega_p^{(k)} \in \mathcal{P}^{(k)}} \;\{\Omega_{p,s}^{(k+1)}\}_{s=1}^S.
\]
This diagnose-and-design evolution operationalizes \(\mathcal{F}\) for producing increasingly adaptive memory systems, ensuring that architectural updates are both empirically grounded and structurally constrained within the unified design space.


\section{Experiments}
\label{sec:exp}

\begin{table}[!t]
\centering
\caption{Performance of various agent frameworks on the WebWalerQA, xBench-Ds, TaskCraft, and GAIA benchmarks.}
\vspace{-0.5em}
\label{tab:main-table-1}
\setlength{\tabcolsep}{6pt}
\resizebox{\textwidth}{!}{
\begin{tabular}{ll|ccccccc}
\toprule
\midrule
\multirow{2}{*}{\textbf{Framework}} & \multirow{2}{*}{\textbf{Model Family}} & \multirow{2}{*}{\textbf{\makecell{WebWalker\\QA}}} & \multirow{2}{*}{\textbf{\makecell{xBench\\-DS}}} & \multirow{2}{*}{\textbf{\makecell{Task\\Craft}}} & \multicolumn{4}{c}{\textbf{GAIA}} \\
\cmidrule(lr){6-9}
& & & & & \textbf{Avg.} & \textbf{Level 1} & \textbf{Level 2} & \textbf{Level 3} \\
\Xhline{0.5pt}
\multicolumn{9}{l}{\cellcolor[HTML]{9ECFD4}{\textit{Closed-source Agent Frameworks}}} \\
\Xhline{0.5pt}
Langfun & Claude 3.7 \etc{} &- & -& - & 71.52  & 83.02 &  68.60 & 57.69 \\
TraseAgent & Claude \etc{} & - & - & - & 70.30 & 83.02 & 69.77 & 46.15 \\
OpenAI Deep Research & o1, o3 \etc{} & - & - & - & 67.36 & 74.29 & 69.06 & 47.60 \\
h2oGPTe & Claude-3.5 & - & - & - & 63.64 & 67.92 & 67.44 & 42.31 \\
Desearch & GPT-4o & - & - & - & 56.97 & 71.70 & 58.14 & 23.08 \\
\Xhline{0.5pt}
\multicolumn{9}{l}{\cellcolor[HTML]{70B2B2}{\textit{Open-Source Agent Frameworks}}} \\
\Xhline{0.5pt}
OWL Workforce (pass@3) & GPT-4o+o3-mini & 57.64 & 55.0 & 58.33  & 60.61 & 81.14 & 58.14 & 26.92 \\
OWL RP (pass@3) & GPT-4o+o3-mini & - & - & - & 58.18 & 81.14 & 54.65 & 23.08 \\
TapeAgents & Claude 3.7 \etc{} & - &-  &-  & 55.76 & 71.70 & 53.49 & 30.77 \\
AutoAgent & Claude 3.5 \etc{} & - & - & - & 55.15 & 71.70 & 53.40 & 26.92 \\
Smolagents \includegraphics[width=0.3cm]{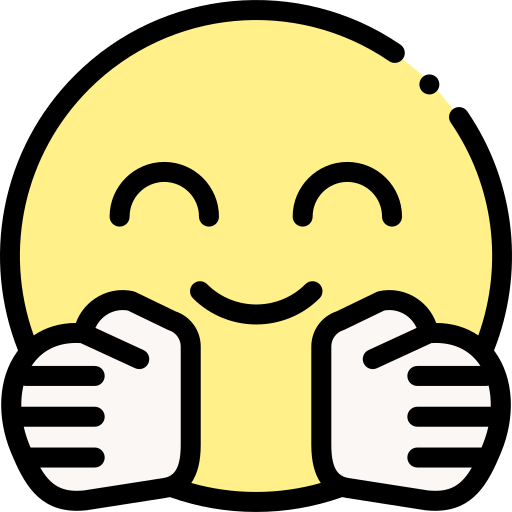} & GPT-4.1 & - & - & - & 55.15 & 67.92 & 53.49 & 34.62 \\
Smolagents \includegraphics[width=0.3cm]{figs/hf.png} & GPT-5-mini & 58.82 & 51.0 & 64.00 & 55.75 & 69.81 & 54.65 & 30.77 \\
Magnetic-1 & OpenAI o1 \etc{} & - & - & - & 46.06 & 56.60 & 46.51 & 23.08 \\
Cognitive Kernel-Pro (pass@1) & Claude-3.7 \etc{} & 60.64 &  56.0 &  66.00  & 60.00 & 79.25 & 56.98 & 30.77 \\
Cognitive Kernel-Pro (pass@3) & Claude-3.7 \etc{} & - & - & - & 75.15 & 84.91 & 73.26 & 61.54 \\
OAgents & Claude-3.7 \etc{} & 58.23 & 47.0 & - & 66.67 & 77.36 & 66.28 & 46.15 \\
JoyAgents & Claude-4, o4-mini & - & - & - & 75.2 & 86.8 & 77.9 & 42.3   \\
Agent KB (pass@1) & GPT-4.1 &  60.59 & 48.0 &61.67  & 61.21 & 79.25 & 58.14 & 34.62 \\
Agent KB (pass@2) & GPT-4.1 &  68.82 &  58.0 &  72.67 & 67.27 & 83.02 & 67.44 & 34.62 \\
Agent KB (pass@3) & GPT-4.1 &  73.53 & 68.0 &  75.33 & 73.94 & 84.91 & 73.26 & 53.85 \\
Flash-Searcher \includegraphics[width=0.3cm]{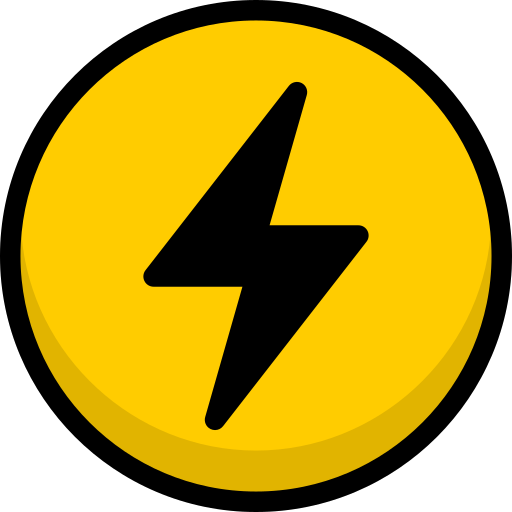} (pass@1) & GPT-5-mini & 71.18 & 69.0 & 69.67 & 69.09 & 79.25 & 69.77 & 46.15 \\
Flash-Searcher \includegraphics[width=0.3cm]{figs/flash.png} (pass@1) & Kimi K2 & 52.35 & 66.0 & 58.00 & 52.12 & 58.49 & 52.33 & 34.62 \\
Flash-Searcher \includegraphics[width=0.3cm]{figs/flash.png} (pass@1) & DeepSeek V3.2 & 69.41 & 68.0 & 69.33 & 60.61 & 79.25 & 53.49 & 46.15 \\

\midrule
\ourmethod+ \includegraphics[width=0.3cm]{figs/hf.png} (pass@1) & GPT-5-mini &  61.18 & 57.0 & 67.67 & 64.24 & 83.02 & 58.14 & 46.15  \\
\ourmethod+ \includegraphics[width=0.3cm]{figs/hf.png} (pass@2) & GPT-5-mini &  67.06 & 63.0 & 75.00 & 67.88 & 84.91 & 63.95 & 46.15  \\
\ourmethod+ \includegraphics[width=0.3cm]{figs/hf.png} (pass@3) & GPT-5-mini &  71.18 & 68.0 & 77.00 & 72.12 &  88.68 & 68.60 & 50.00 \\
\midrule
\ourmethod+ \includegraphics[width=0.3cm]{figs/flash.png}  (pass@1) & GPT-5-mini & 74.71 & 74.0 & 72.00 & 73.33 & 83.02 & 73.26 & 53.85 \\
\ourmethod+ \includegraphics[width=0.3cm]{figs/flash.png} (pass@2) & GPT-5-mini & 79.41 & 77.0 & 75.00 & 77.58 & 92.45 & 74.42 & 57.69  \\
\ourmethod+ \includegraphics[width=0.3cm]{figs/flash.png} (pass@3) & GPT-5-mini & 81.18 & 78.0 & 79.33 & 80.61 & 94.34 & 79.07 & 57.69\\

\midrule
\ourmethod+ \includegraphics[width=0.3cm]{figs/flash.png} (pass@1) & Kimi K2 & 69.41 & 68.0 & 68.00 &  61.21 & 67.92 & 63.95 & 38.46 \\
\ourmethod+ \includegraphics[width=0.3cm]{figs/flash.png} (pass@1) & DeepSeek V3.2 & 72.35 & 70.0 & 72.67 & 67.88 & 83.02 & 63.95 & 50.00   \\
\midrule
\bottomrule
\end{tabular}}
\vspace{-1em}
\end{table}

\subsection{Experiment Setup}
\vspace{-0.7em}

\fakeparagraph{Benchmarks} We evaluate the proposed framework across four challenging agentic benchmarks, including
\textit{GAIA}~\citep{mialon2023gaia}, \textit{WebWalkerQA}~\citep{wu2025webwalkerbenchmarkingllmsweb}, \textit{xBench-DeepSearch} (xBench-DS)~\citep{chen2025xbenchtrackingagentsproductivity}, as well as 
\textit{TaskCraft}~\citep{shi2025taskcraft}. 
Further statistics and details are provided in \Cref{app:dataset}.

\vspace{-0.5em}
\fakeparagraph{Method Configurations}
We run the dual-evolution process for \(K_{\max}=3\) iterations. In the outer loop, the survivor budget is set as \(K=1\); at each iteration, only the top-ranked architecture is retained and expanded to \(S=3\) descendants. 
In the inner loop, each candidate architecture \(\Omega_j^{(k)}\) is evaluated on a batch \(\mathcal{T}_j^{(k)}\) of 60 task trajectories, consisting of 40 newly sampled tasks and 20 tasks reused from the previous iteration to stabilize inter-iteration comparison.

\vspace{-0.5em}
\fakeparagraph{Agent Framework} We integrate \ourmethod into two representative agentic frameworks: 
\includegraphics[width=0.3cm]{figs/hf.png} \textit{SmolAgent}~\citep{smolagents}, a lightweight two-agent architecture, and 
\includegraphics[height=0.8\baselineskip]{figs/flash.png} \textit{Flash-Searcher}~\citep{qin2025flashsearcherfasteffectiveweb}, a high-performance single-agent deep research system. 
To assess the generalization and plug-and-play capability of \ourmethod, we further evaluate it on two held-out multi-agent systems: 
Tencent's \includegraphics[width=0.3cm]{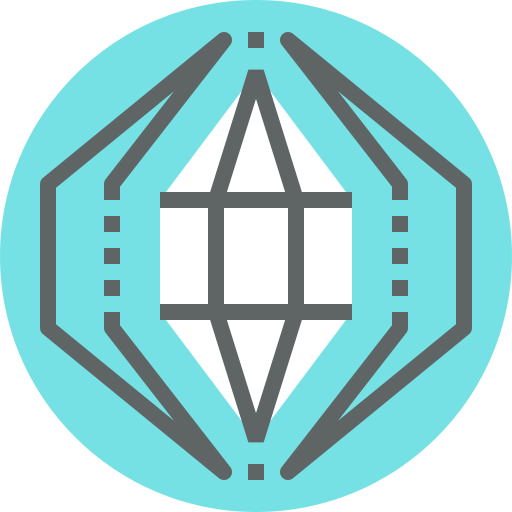} \textit{Cognitive Kernel-Pro (CK-Pro)}~\citep{fang2025cognitivekernelproframeworkdeep}, a three-agent framework comprising main/file/web agents; and 
\includegraphics[height=0.8\baselineskip]{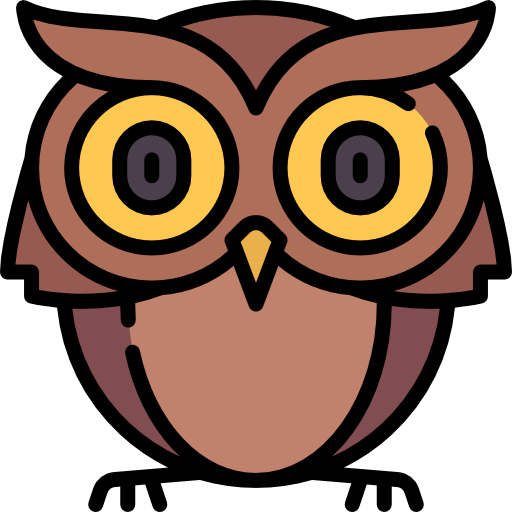} \textit{OWL}~\citep{owl2025}, a hierarchical system including planner, coordinator, web, document, and coding agents. 
This diversity in architecture and system complexity enables a comprehensive examination of the adaptability of \ourmethod across heterogeneous agentic scaffolds.

\vspace{-0.5em}
\fakeparagraph{Model Configurations} We instantiate \ourmethod using \textsc{GPT-5-mini}~\citep{openaiIntroducingGPT5} as the LLM backbone for the underlying agentic frameworks, and for supporting the meta-evolution operator \(\mathcal{F}(\cdot)\). 
To further evaluate the cross-LLM generalization capability of \ourmethod, we additionally consider alternative backbones, including \textsc{DeepSeek V3.2}~\citep{deepseekai2025deepseekv32pushingfrontieropen}, and \textsc{Kimi K2}~\citep{kimiteam2025kimik2openagentic}. 
For clarity, we explicitly report the specific LLM backbone used by each agentic framework in the following experiments.

\subsection{Main Results}
\begin{wrapfigure}{r}{0.5\textwidth}
\vspace{-2em}
  \begin{center}
    \includegraphics[width=0.5\textwidth]{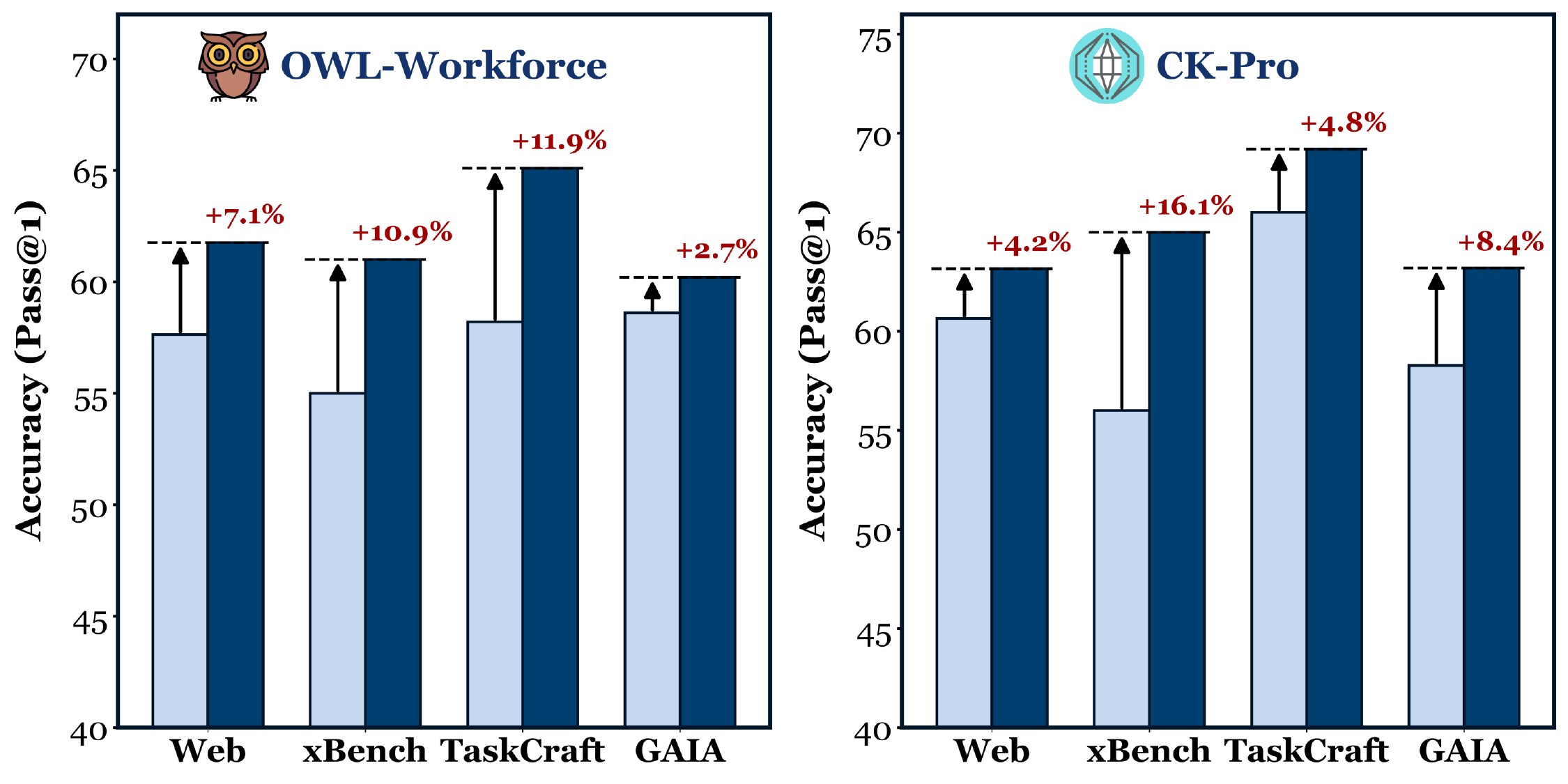}
  \end{center}
  \vspace{-0.8em}
  \caption{The cross-framework generalization analysis. We transfer the memory system evolved on TaskCraft+\includegraphics[height=0.8\baselineskip]{figs/flash.png} to \includegraphics[height=0.8\baselineskip]{figs/owl.png} and \includegraphics[width=0.3cm]{figs/kernel.png}. Red percentages denote the relative score gains of each framework after integrating \ourmethod over its memory-free counterpart.}
  \label{fig:cross-framework}
  \vspace{-2.em}
\end{wrapfigure}
We report the pass@1–3 performance of \ourmethod integrated with {SmolAgent} and {Flash-Searcher} in \Cref{tab:main-table-1}, together with its generalization results when paired with unseen LLMs (\textsc{Kimi K2}, \textsc{DeepSeek V3.2}). 
Notably, on the relatively simple {TaskCraft} benchmark, we evolve two distinct memory systems using MemEvolve+\includegraphics[width=0.3cm]{figs/hf.png} and MemEvolve+\includegraphics[height=0.8\baselineskip]{figs/flash.png}, respectively. These evolved memory systems are then fixed and evaluated on {WebWalkerQA} and {xBench-DS}, \textit{i.e.}, without conducting dataset-specific meta-evolution.

\begin{table}[!t]
\centering
\caption{Performance, cost, delay, and steps across datasets under different memory settings for \includegraphics[width=0.3cm]{figs/flash.png} Flash-Searcher. Here, \textit{cost} denotes the average API cost incurred per task query, \textit{delay} measures the average execution latency (seconds) per task, and \textit{\#steps} reports the number of agent interaction steps required to complete each task.
}
\vspace{-0.5em}
\label{tab:compare-self-evolving-memory}
\small
\setlength{\tabcolsep}{5pt}
\resizebox{\textwidth}{!}{
\begin{tabular}{lcccccccccccc}
\toprule
\midrule
\multirow{2}{*}{\textbf{Memory Setting}} 
& \multicolumn{4}{c}{\textbf{GAIA}} 
& \multicolumn{4}{c}{\textbf{xBench}} 
& \multicolumn{4}{c}{\textbf{WebWalkerQA}} \\
\cmidrule(lr){2-5}\cmidrule(lr){6-9}\cmidrule(lr){10-13}
& {Perf.} & {Cost} & {Delay} & {\#Steps}
& {Perf.} & {Cost} & {Delay} & {\#Steps}
& {Perf.} & {Cost} & {Delay} & {\#Steps} \\
\midrule
No-Memory    & 69.09 & 0.086 & 505.46 & 10.44 & 69.00 & 0.141 & 523.05 & 14.69 & 71.18 & 0.048 & 251.57 & 6.91 \\
Generative   & 66.67 & 0.061 & 436.26 & 8.87  & 70.00 & 0.131 & 818.37 & 13.45 & 72.35 & 0.045 & 268.56 & 6.64 \\
Voyager      & 69.70 & 0.060 & 499.89 & 9.25  & 68.00 & 0.117 & 553.46 & 12.71 & 73.53 & 0.049 & 333.69 & 6.99 \\
DILU         & 66.67 & 0.059 & 444.62 & 8.91  & 69.00 & 0.134 & 500.72 & 13.83 & 72.94 & 0.046 & 272.16 & 6.96 \\
ExpeL        & 66.06 & 0.059 & 500.11 & 8.68  & 64.00 & 0.123 & 710.32 & 13.05 & 69.41 & 0.076 & 385.28 & 10.96 \\

AWM          & 67.27 & 0.062 & 584.88 & 10.23 & 71.00 & 0.138  & 761.33  & 14.12  & 72.35  & 0.068 & 397.20  & 11.40 \\

Mobile-E     & 69.09 & 0.065 & 321.80 & 9.35  & 68.00  & 0.120  &  537.18   &  13.16  & 71.76  & 0.059  & 296.01  &  6.52 \\
Cheatsheet   & 68.48 & 0.069 & 559.81 & 9.72  & 65.00 & 0.174 & 818.07 & 15.99 & 72.94 & 0.057 & 367.13 & 7.59 \\
\ourmethod   & 73.33 & 0.085 & 693.33 & 10.14 & 74.00 & 0.136  &   773.06  &  14.20     & 74.71 & 0.040 & 332.49 & 6.64 \\
\midrule
\bottomrule
\end{tabular}
}
\vspace{-0.6em}
\end{table}

\fakeparagraph{Memory System Matters For Agent Systems} As shown in \Cref{tab:main-table-1}, equipping agentic systems with effective memory architectures is critical to performance. On xBench, \includegraphics[width=0.3cm]{figs/hf.png}+\textsc{GPT-5-Mini} achieves an initial pass@1 of $51\%$; after integrating \ourmethod, pass@1 increases by $6\%$, while pass@3 goes up to $68.0\%$. Similarly, \includegraphics[height=0.8\baselineskip]{figs/flash.png}+\textsc{GPT-5-Mini} improves from $69\%$ to $74\%$ on xBench when augmented with \ourmethod. These results clearly demonstrate the substantial impact of a well-designed memory system on agent performance.
At the same time, memory is not a panacea and remains bounded by the capabilities of the underlying agentic framework. On GAIA, \ourmethod+\includegraphics[width=0.3cm]{figs/hf.png} attains a pass@3 of $72.12\%$, comparable to AgentKB, while avoiding the construction of large and costly offline knowledge bases. In contrast, the gains with \ourmethod+\includegraphics[width=0.3cm]{figs/flash.png} are even more pronounced, achieving a pass@3 of $80.61\%$, surpassing several strong multi-agent systems such as OWL-Workforce and CK-Pro under the same metric.

\fakeparagraph{\ourmethod Exhibits Cross-Task, Cross-Model, and Cross-Framework Generalization}
Recall that the memory systems used on WebWalkerQA and xBench are directly inherited from those evolved on TaskCraft, without any task-specific meta-evolution. Nevertheless, these transferred memories yield consistent gains on more challenging benchmarks (WebWalkerQA+\includegraphics[width=0.3cm]{figs/hf.png}: $58.82 \rightarrow 61.18\%$; xBench+\includegraphics[height=0.8\baselineskip]{figs/flash.png}: $69.0 \rightarrow 74.0\%$), indicating that \ourmethod captures task-agnostic principles of memory design rather than overfitting to individual datasets.
\ourmethod also demonstrates strong cross-LLM generalization. Although meta-evolution is conducted using \textsc{GPT-5-Mini}, memory systems evolved on TaskCraft+\includegraphics[height=0.8\baselineskip]{figs/flash.png} transfer effectively to \textsc{Kimi K2} and \textsc{DeepSeek V3.2} \emph{without manual adaptation}. Notably, \textsc{Kimi K2}+\includegraphics[height=0.8\baselineskip]{figs/flash.png} improves by $17.06\%$ on WebWalkerQA and $10.0\%$ on TaskCraft.
Finally, \ourmethod exhibits compelling cross-framework generalization. As shown in \Cref{fig:cross-framework}, directly transferring the memory system evolved on TaskCraft+\includegraphics[height=0.8\baselineskip]{figs/flash.png} to heterogeneous agentic frameworks, including \includegraphics[height=0.8\baselineskip]{figs/owl.png} and \includegraphics[width=0.3cm]{figs/kernel.png}, consistently improves performance despite substantial architectural differences. These results demonstrate that \ourmethod learns framework-agnostic memory abstractions that are readily pluggable across diverse agentic systems.

\begin{figure}[!t]
    \centering
    \includegraphics[width=\linewidth]{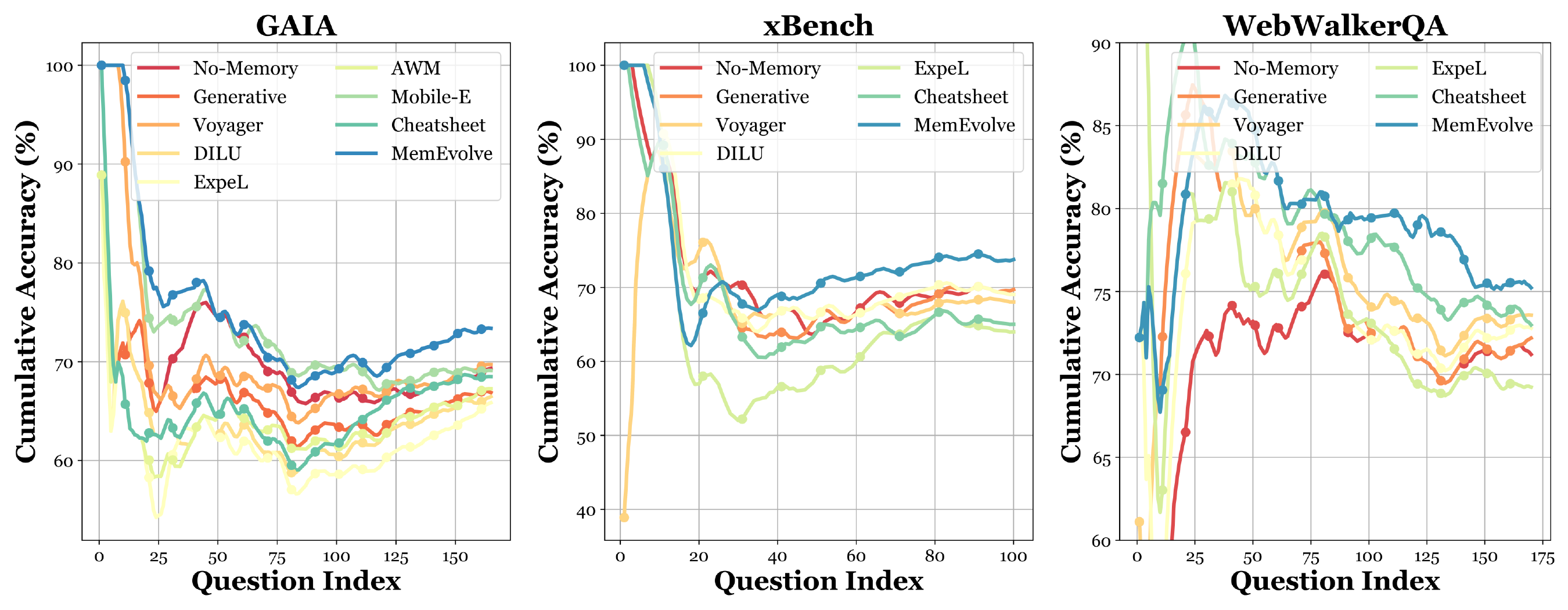}
    \vspace{-2em}
    \caption{Evolution of cumulative accuracy across question indices. Cumulative accuracy at index $i$ is defined as the average accuracy over the first $i$ questions. The curves exhibit larger fluctuations at early indices due to limited sample size, and gradually stabilize as more questions are accumulated.}
    \label{fig:curve}
    \vspace{-0.6em}
\end{figure}

\begin{figure}[!h]
    \centering
    \includegraphics[width=\linewidth]{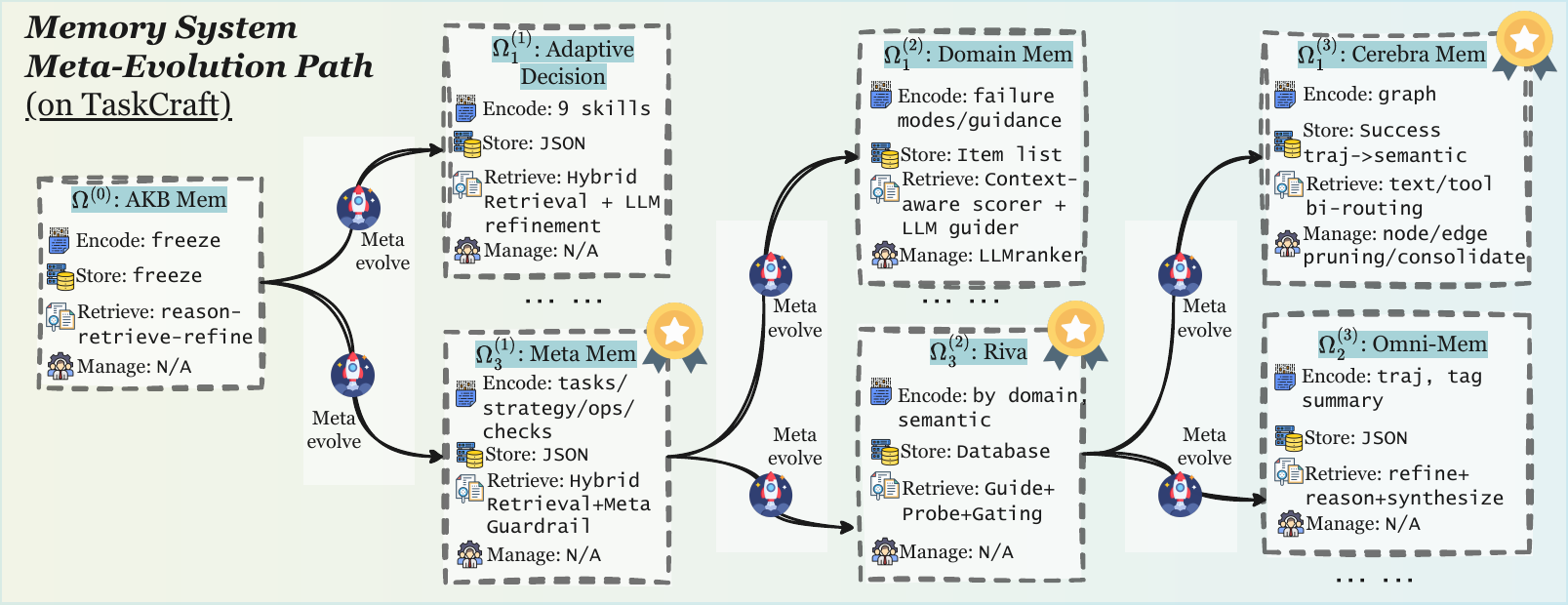}
    \vspace{-1.6em}
    \caption{Illustration of the progressive evolution from the fixed AgentKB architecture to increasingly agentic and efficient memory architectures. Each stage reflects structural and functional modifications in memory encoding, storing, retrieval, and maintenance, culminating in high-performing systems such as Riva and Cerebra.}
    \label{fig:path1}
    \vspace{-0.5em}
\end{figure}

\subsection{Self-Evolving Memory Comparison}

We further compare the memory systems automatically evolved by \ourmethod against prevailing human-designed self-improving memory systems. In \Cref{tab:compare-self-evolving-memory}, we integrate seven representative self-improving memory systems implemented in \ourframework with Flash-Searcher, and comprehensively report performance, per-task cost/execution latency/execution steps. Results for \ourmethod are obtained using the system evolved on TaskCraft+\includegraphics[height=0.8\baselineskip]{figs/flash.png}+\textsc{GPT-5-Mini}.

\fakeparagraph{Existing Memory Systems Fail to Deliver Consistent Gains}
Despite faithful re-implementations aligned with the original designs, many existing memory systems do not yield stable improvements. For example, DILU improves performance on xBench and WebWalkerQA, yet degrades GAIA by $2.42\%$. Dynamic Cheatsheet achieves a $1.76\%$ gain on WebWalkerQA via skill condensation, but performs poorly on GAIA and xBench. More extreme cases are also observed: ExpeL underperforms on all three benchmarks. Upon closer inspection, this is unsurprising, as ExpeL was originally designed for relatively simple embodied or QA settings (\textit{e.g.}, ALFWorld, HotpotQA), and its prompts and mechanisms are ill-suited for long-horizon, long-context deep research. These results underscore the necessity of task-aware memory design.

\fakeparagraph{\ourmethod Delivers Robust and Consistent Improvements}
In contrast to prior approaches, \ourmethod yields stable and robust performance gains. Although the underlying memory system is evolved on TaskCraft, it consistently achieves improvements of $3.54\%\sim5.0\%$ across all three evaluated benchmarks. Importantly, these gains are not achieved by substantially increasing the per-task cost. As shown in \Cref{tab:compare-self-evolving-memory}, \ourmethod maintains API costs comparable to the No-Memory baseline across all benchmarks (\textit{e.g.}, GAIA: \$0.085 vs. \$0.086; xBench: \$0.136 vs. \$0.141), while its execution delay remains on a similar scale to other self-improving baselines (\textit{e.g.}, GAIA: $693.33$s vs. $584.88$s for AWM and $559.81$s for Cheatsheet; xBench: $773.06$s vs. $761.33$s for AWM and $818.07$s for Cheatsheet). \Cref{fig:curve} further illustrates the \textit{cumulative success rate} of different self-evolving memory systems as task execution progresses. Although performance exhibits higher variance in the early stages due to limited sample size, \ourmethod gradually stabilizes and converges to a consistently superior performance regime.
 This indicates that \ourmethod discovers principled and effective memory designs rather than relying on brittle, task-specific heuristics. 
 
At first glance, such generalization may appear to conflict with our original motivation that memory systems cannot generalize across all domains and therefore require task-specific evolution. We argue this is not the case. Memory systems evolved on TaskCraft are unlikely to transfer effectively to fundamentally different task families (\textit{e.g.}, embodied action), where environments, action space and tool sets differ substantially. Nevertheless, \ourmethod enables the discovery of broadly applicable memory architectures within a shared task regime, while retaining the capacity for further task-specific adaptation when required.

\begin{figure}[!t]
    \centering
    \includegraphics[width=\linewidth]{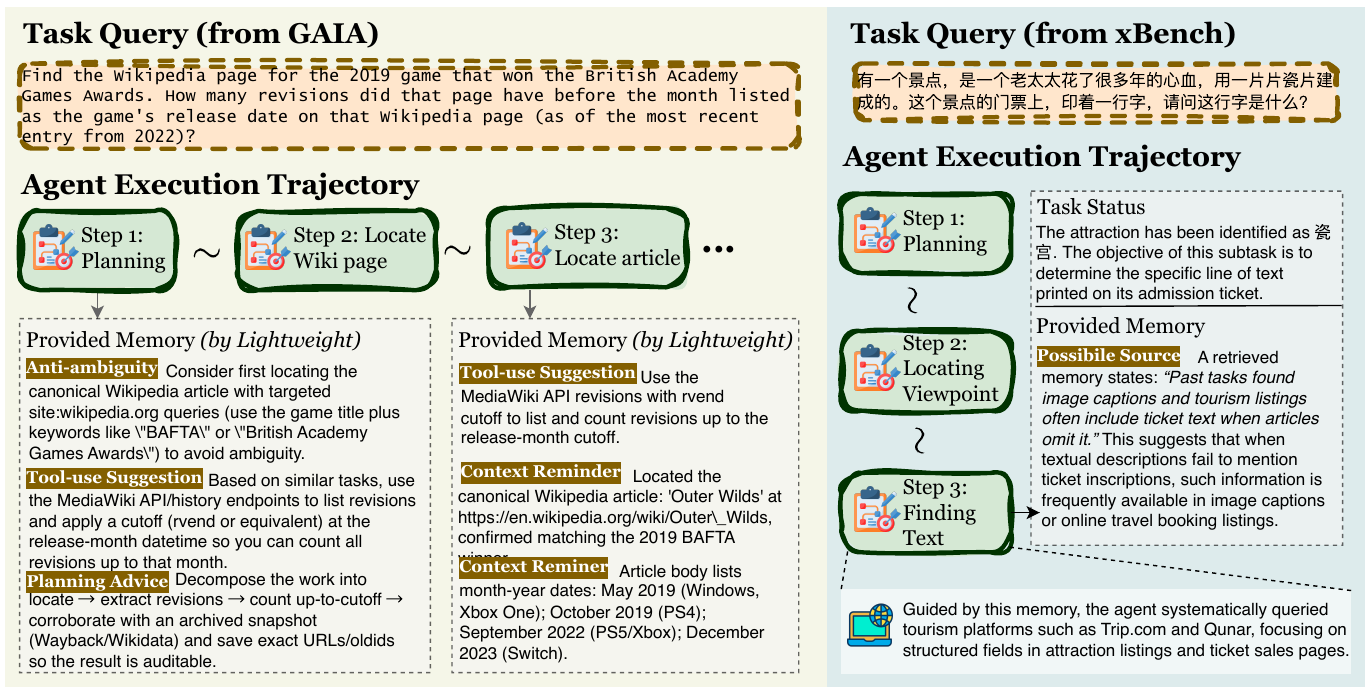}
    \vspace{-1.5em}
    \caption{Illustration of how evolved memories are instantiated during real-world tasks from GAIA and xBench. The memory system adaptively provides stage-specific guidance, ranging from high-level planning and task decomposition to fine-grained tool-use suggestions and salient context recall, thereby steering the agent toward efficient and successful task completion.}
    \label{fig:case}
    \vspace{-0.5em}
\end{figure}

\subsection{Meta-Evolving Dynamics}

Having established the substantial performance gains delivered by \ourmethod, we further examine \emph{how} meta-evolution is executed in practice and \emph{which} components are modified or enhanced during the evolutionary process. As illustrated in \Cref{fig:path1}, \ourmethod starts from the predefined structure of AgentKB and iteratively evolves toward increasingly efficient memory architectures. \Cref{fig:cerebra,fig:riva} highlights two high-performing memory systems discovered along this trajectory, denoted as \textsc{Riva} and \textsc{Cerebra}.  \Cref{fig:lightweight} presents a system evolved from the simplest few-shot example memory baseline, referred to as \textsc{Lightweight}.

\fakeparagraph{Agents Spontaneously Evolve Efficient Memory Architectures}
As illustrated in \Cref{fig:path1}, the initial AgentKB memory system adopts a frozen design for both encoding and storage, lacking the capability to assimilate new experiences. Starting from this baseline, \ourmethod explores a spectrum of evolutionary directions. Some candidates are relatively aggressive (e.g., $\Omega^{(1)}_1$, an \textsc{Adaptive Decision System} that decomposes a single agent trajectory into nine skill granularities), while others are more conservative (e.g., $\Omega^{(1)}_3$, an \textsc{Meta Memory System} that stores trajectories at four levels and introduces an LLM-based meta-guardrail during retrieval to filter irrelevant information). The latter emerges as the winner in the first evolutionary round.
The defining characteristic of this stage is \textit{agentic}: both memory encoding and decoding increasingly rely on agent-driven decisions rather than predefined pipelines. The third evolution round introduces two further advances. Evolving from $\Omega^{(2)}_3$ \textsc{Riva} to $\Omega^{(3)}_1$ \textsc{Cerebra}, the memory system learns to distill not only textual insights but also reusable tools from past experience, while incorporating periodic maintenance of the memory database. Together, these enhancements provide faster evolutionary momentum for underlying agentic frameworks.

\fakeparagraph{Evolved Memory Systems Are Effective in Practice}
We further present concrete memory examples produced by the \textsc{Lightweight} system during real executions, as shown in \Cref{fig:case}. The results illustrate that \textsc{Lightweight} delivers memory content at varying levels of granularity, adaptively tailored to different task stages. During early planning, the memory provides high-level guidance, such as task decomposition strategies. As execution proceeds, it offers more fine-grained recommendations for tool-use, along with a form of \textit{working memory} that highlights salient information from previous turns. Notably, \textsc{Lightweight} also exhibits predictive behavior by anticipating that target information may appear within image content on online travel websites, successfully guiding the agent to locate the evidence on trip.com. Together, these examples demonstrate the practical effectiveness of memory systems evolved by \ourmethod.

\section{Conclusion}
\label{sec:conclusion}

This work provides a unified implementation and design space for the rapidly growing field of self-evolving agent memory, together with a standardized codebase, termed \ourframework, upon which we further build \ourmethod, a meta-evolutionary memory framework. Departing from the conventional paradigm of manually crafting a single self-improving memory architecture and expecting it to generalize across all domains, \ourmethod instead embraces adaptive, architecture-level evolution driven by empirical interaction feedback. Extensive experiments across diverse agentic benchmarks and backbone models demonstrate the effectiveness, robustness, and generalization of this approach. Moreover, analysis of the automatically evolved memory systems reveals several instructive design principles, including increased agentic involvement, hierarchical organization, and multi-level abstraction. We hope that \ourmethod serves as a step toward more automated, principled, and meta-evolutionary pathways for building continually improving agentic intelligence.

\input{acknowledge}
\clearpage
\bibliographystyle{apalike}

\bibliography{cite.bib}

\clearpage
\beginappendix

\section{EvolveLab Implementation}\label{app:EvolveLab}

\ourframework is designed as a modular and extensible codebase to support the systematic study of self-evolving agent memory systems. It provides a unified interface that abstracts the complexities of diverse memory architectures, enabling standardized implementation, evaluation, and meta-evolution.

\subsection{Unified Interface and Abstract Base Class}
The cornerstone of \ourframework{} is the \texttt{BaseMemoryProvider} abstract base class (ABC), which defines the fundamental protocol for all memory systems. As shown in the code snippet below, the interface enforces two primary operations that map to the modular design space (\textit{Encode, Store, Retrieve, Manage}):

\begin{itemize}[leftmargin=1.5em, itemsep=0pt]
    \item \textbf{Retrieve (\texttt{provide\_memory})}: Handles context-aware memory recall. It accepts a \texttt{MemoryRequest} containing the current task query, execution context, and system status, and returns a \texttt{MemoryResponse} containing a list of relevant \texttt{MemoryItem}s.
    \item \textbf{Encode \& Store (\texttt{take\_in\_memory})}: Orchestrates the ingestion of new experiences. This method processes a \texttt{TrajectoryData} object, which encapsulates the complete history of a task execution, extracts structural insights or tools (\textit{Encode}), and persists them into the underlying storage medium (\textit{Store}).
\end{itemize}

While \texttt{take\_in\_memory} primarily integrates the \textit{Encode} and \textit{Store} stages, the \textit{Manage} functionality that is responsible for offline consolidation or selective forgetting is typically implemented as auxiliary methods within the provider classes or invoked during specific lifecycle events.

\begin{lstlisting}[language=Python, caption=The Abstract Base Class of Memory Providers]
class BaseMemoryProvider(ABC):
    """Abstract base class for memory providers"""
    
    def __init__(self, memory_type: MemoryType, config: Optional[dict] = None):
        self.memory_type = memory_type
        self.config = config or {}
    
    @abstractmethod
    def provide_memory(self, request: MemoryRequest) -> MemoryResponse:
        """
        Retrieve relevant memories based on query, context and status
        Args:
            request: MemoryRequest containing query, context, status and optional params
        Returns:
            MemoryResponse containing relevant memories
        """
        pass
    
    @abstractmethod
    def take_in_memory(self, trajectory_data: TrajectoryData) -> tuple[bool, str]:
        """
        Store/ingest new memory from trajectory data
        Args:
            trajectory_data: TrajectoryData containing query, trajectory and metadata
        Returns:
            tuple[bool, str]: (Success status of memory ingestion, Description of absorbed memory)
        """
        pass
    
    @abstractmethod
    def initialize(self) -> bool:
        """
        Initialize the memory provider (load existing data, setup indices, etc.)
        Returns:
            bool: Success status of initialization
        """
        pass
    
    def get_memory_type(self) -> MemoryType:
        """Get the type of this memory provider"""
        return self.memory_type
    
    def get_config(self) -> dict:
        """Get the configuration of this memory provider"""
        return self.config.copy()
\end{lstlisting}

\subsection{Standardized Data Carriers}
To ensure seamless interoperability across heterogeneous memory designs and agent frameworks, \ourframework utilizes standardized memory data carriers. These structures act as the "universal language" of the framework:

\begin{itemize}[leftmargin=1.5em, itemsep=0pt]
    \item \textbf{\texttt{MemoryItem}}: The fundamental unit of information, capable of representing raw text, distilled insights, or executable code (APIs). Each item includes metadata such as creation timestamps, confidence scores, and source identifiers.
    \item \textbf{\texttt{TrajectoryData}}: A comprehensive container for task execution history, including the initial query, full interaction traces (state-action pairs), and terminal rewards. It serves as the raw substrate for memory evolution.
    \item \textbf{\texttt{MemoryRequest/Response}}: Standardized envelopes for retrieval queries and results, ensuring that any agent system can interact with any memory provider without architecture-specific modifications.
\end{itemize}

\subsection{Implementation Examples: ExpeL and SkillWeaver}
The versatility of the \ourframework interface is demonstrated by our implementation of twelve distinct memory systems. Two representative examples are:

\begin{itemize}[leftmargin=1.5em, itemsep=2pt]
    \item \textbf{ExpeLProvider}: Implements a contrastive learning-based memory. Its \texttt{take\_in\_memory} function identifies successful and failed trajectories to distill high-level "insights" into a textual format. These insights are stored in a vector database and retrieved via semantic similarity during \texttt{provide\_memory} to guide the agent away from previous mistakes.
    \item \textbf{SkillWeaverProvider}: Operates in a tool-centric design space. Its \texttt{take\_in\_memory} logic uses an LLM to synthesize reusable Python functions (skills) from successful trajectories. These skills are stored as executable code-level repositories and are dynamically retrieved and injected into the agent's action space through the unified \texttt{MemoryItem} interface.
\end{itemize}

\section{Experiment Details}
\subsection{Dataset Details}\label{app:dataset}
The four datasets used in this study are described and summarized as follows:
\begin{itemize}
\item \textit{GAIA}~\citep{mialon2023gaia} consists of 165 tasks, including 53 Level-1, 86 Level-2, and 26 Level-3 problems. For evaluating \ourmethod on GAIA+\includegraphics[width=0.3cm]{figs/hf.png} and GAIA+\includegraphics[width=0.3cm]{figs/flash.png}, the memory systems are evolved using GAIA Level-1 tasks together with 67 TaskCraft queries. Meta-evolution is conducted for three rounds, with 40 trajectories per round.
\item \textit{WebWalkerQA}~\citep{wu2025webwalkerbenchmarkingllmsweb} evaluates an agent’s ability to handle complex, multi-turn web interactions, comprising 680 real-world queries across four domains and over 1,373 webpages. We sample a subset of 170 queries for evaluation, with the sampling script released in our codebase. All memory systems used for WebWalkerQA are meta-evolved on TaskCraft.
\item \textit{xBench-DeepSearch} (xBench-DS)~\citep{chen2025xbenchtrackingagentsproductivity} contains 100 tasks that assess agentic planning, tool use, and reasoning. Similar to WebWalkerQA, the memory systems used for xBench-DS evaluation are entirely meta-evolved on TaskCraft.
\item \textit{TaskCraft}~\citep{shi2025taskcraft} is a synthetic benchmark generated via an autonomous data pipeline. We collect 300 queries as a working subset and use 120 of them for three rounds of meta-evolution, with 40 queries per round. Meta-evolution for \includegraphics[width=0.3cm]{figs/hf.png} and \includegraphics[width=0.3cm]{figs/flash.png} is performed independently.
\end{itemize}

\section{Memory System Demonstration}
\begin{figure}[!h]
    \centering
    \includegraphics[angle=90,width=0.8\linewidth]{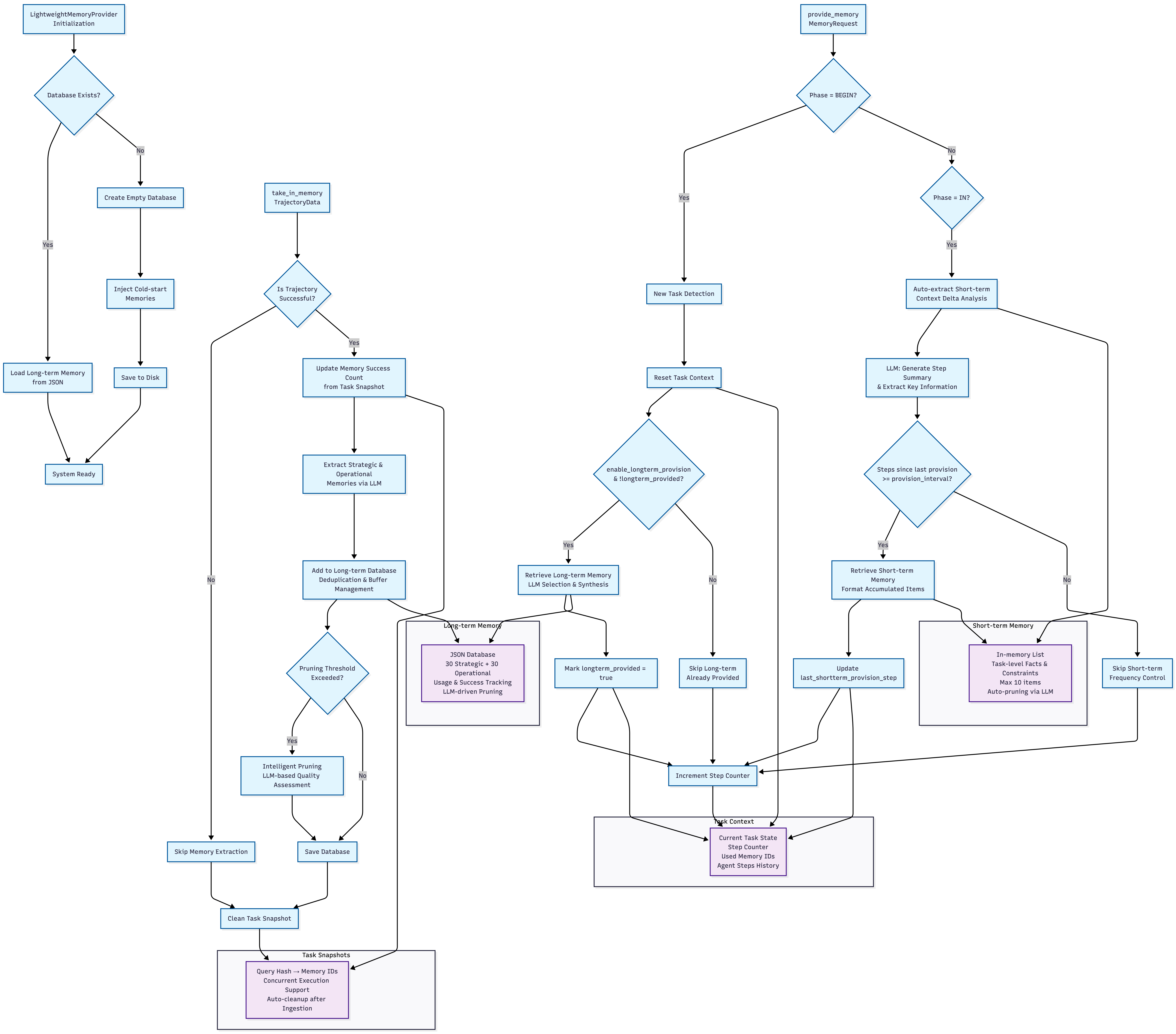}
\caption{Illustration of the \textsc{Lightweight} memory system evolved by \ourmethod. 
The evolutionary starting point is a minimal few-shot trajectory memory, similar to \textsc{MemoryBank}, where each completed trajectory is stored verbatim. For a new task, the agent retrieves the top-$k$ most similar trajectories via vector similarity and directly conditions on them. \ourmethod progressively refines this baseline into a more structured and stage-aware memory system.}

    \label{fig:lightweight}
\end{figure}

\begin{figure}[!h]
    \centering
    \includegraphics[width=\linewidth]{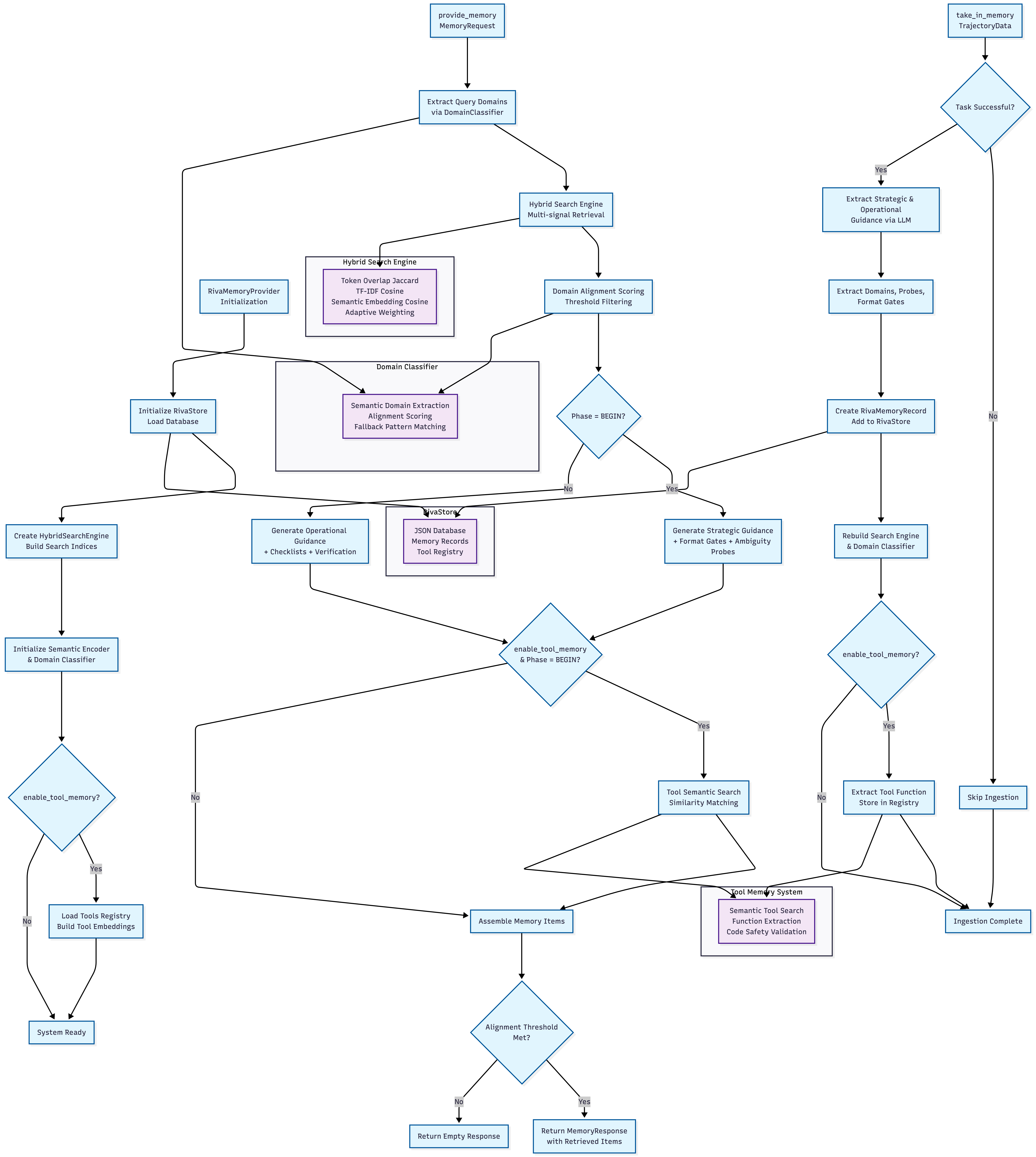}
\caption{Illustration of the \textsc{Riva} memory system evolved by \ourmethod. 
Its evolutionary initialization follows an \textsc{AgentKB}-style architecture, but without inheriting the large and costly offline knowledge base. Through meta-evolution, \textsc{Riva} develops more agent-centric encoding and retrieval strategies while remaining lightweight and fully online.}

    \label{fig:riva}
\end{figure}

\begin{figure}[!h]
    \centering
    \includegraphics[width=0.9\linewidth]{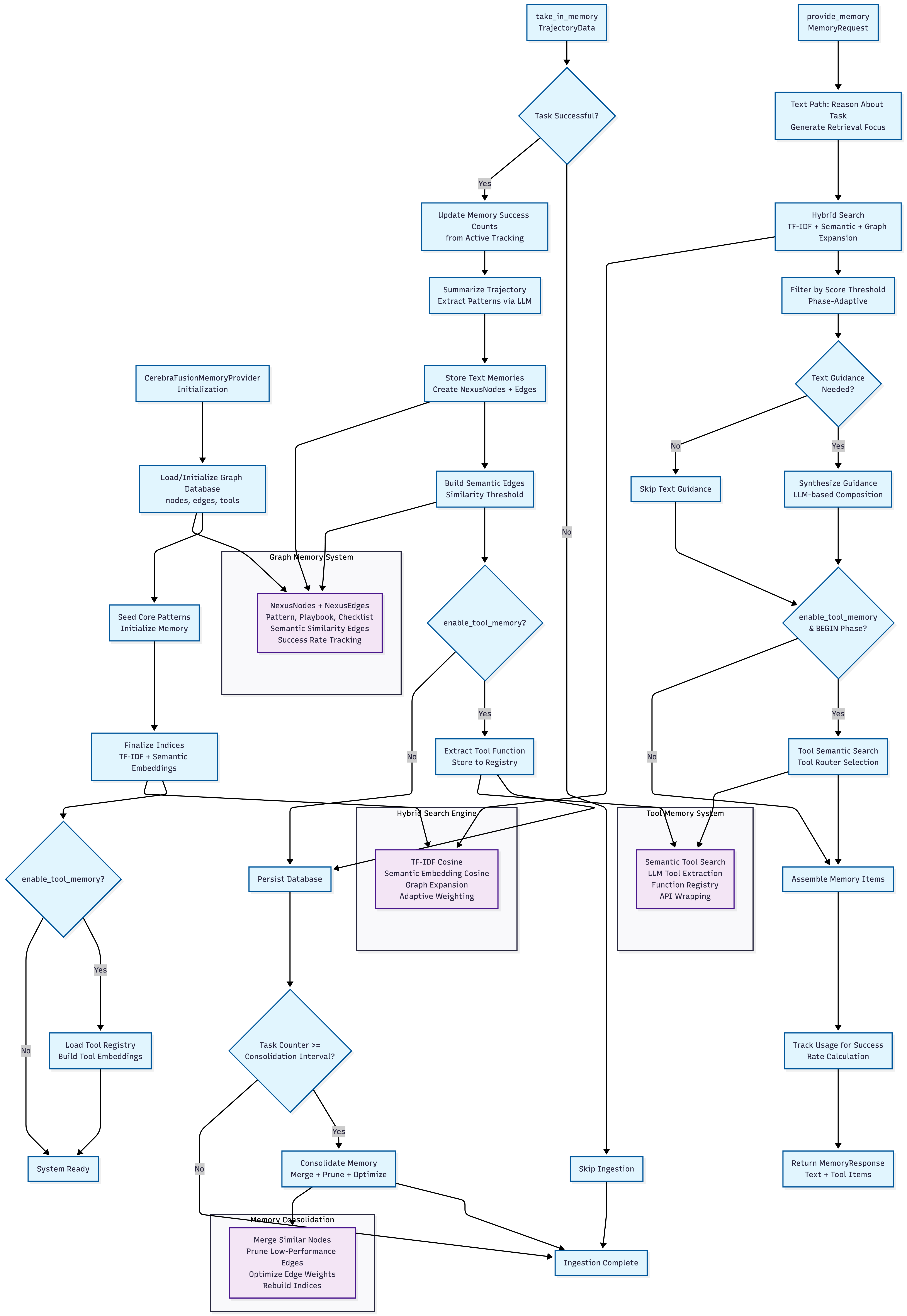}
\caption{Illustration of the \textsc{Cerebra} memory system evolved by \ourmethod. 
Starting from the same \textsc{AgentKB}-style initialization (without the offline knowledge base), \textsc{Cerebra} further evolves to distill both reusable tools and abstract knowledge from experience, and incorporates working memory maintenance mechanisms to support long-horizon agent evolution.}

    \label{fig:cerebra}
\end{figure}

To provide a concrete and intuitive understanding of the memory architectures evolved by \ourmethod, we visualize three representative systems discovered along different evolutionary trajectories, as shown in \Cref{fig:lightweight,fig:riva,fig:cerebra}. These examples highlight how \ourmethod progressively transforms simple, static memory mechanisms into more expressive and adaptive architectures by modifying memory encoding, retrieval, and management strategies. Together, they illustrate the diversity of memory designs that can emerge under the same meta-evolutionary framework.

\end{document}

%% file: common.tex
\usepackage{latexsym}
\usepackage[T1]{fontenc}
\usepackage[utf8]{inputenc}
\usepackage{microtype}
\usepackage{inconsolata}
\usepackage{graphicx}
\usepackage{hyperref}       
\usepackage{url}            
\usepackage{booktabs}       
\usepackage{amsfonts}       
\usepackage{nicefrac}       
\usepackage{stackengine}
\usepackage{microtype}      
\usepackage{colortbl}
\usepackage{xcolor}
\usepackage{amsmath}
\usepackage{amssymb}
\usepackage{amsthm}
\usepackage{mathrsfs}
\usepackage{pifont}
\usepackage{MnSymbol}
\usepackage{balance}
\usepackage{enumitem}
\usepackage{listings}
\usepackage{xcolor}
\usepackage{natbib}
\usepackage{multicol}

\AtBeginDocument{%
  \providecommand\BibTeX{{%
    \normalfont B\kern-0.5em{\scshape i\kern-0.25em b}\kern-0.8em\TeX}}}

\makeatletter
\DeclareRobustCommand\onedot{\futurelet\@let@token\@onedot}
\def\@onedot{\ifx\@let@token.\else.\null\fi}

\newcommand{\etc}{\emph{etc\@\onedot}}

\usepackage{setspace}
\usepackage{mathtools}

\usepackage{multirow,booktabs}
\usepackage{subcaption}

\newcommand{\owo}[1]{\textsc{OAgents}}

\definecolor{lightgreen}{RGB}{144, 238, 144} 
\definecolor{lightred}{RGB}{255, 105, 97}

\newcommand{\fakeparagraph}[1]{\vspace{1mm}\noindent\textbf{#1.}}

\newtcolorbox{promptbox}[2][Prompt]{
colback=black!5!white,
arc=5pt, 
boxrule=0.5pt,
fonttitle=\bfseries,
title=#1, 
before upper={\small}, fontupper=\fontfamily{ptm}\selectfont,
colframe=#2, 
}
\definecolor{ogreen}{RGB}{34, 139, 34}

%% file: acknowledge.tex
\clearpage
\section{Contributions}

\begin{multicols}{2}
\textbf{Core Contributors}
\begin{itemize}
    \item Guibin Zhang
    \item Haotian Ren
\end{itemize}
\textbf{Contributors}
\begin{itemize}
    \item Chong Zhan
    \item Zhenhong Zhou
    \item Junhao Wang
    \item He Zhu
\end{itemize}

\textbf{Corresponding Authors}
\begin{itemize}
\item Wangchunshu Zhou
\item Shuicheng Yan
\item[]
\item[]
\item[]
\item[]
\end{itemize}
\end{multicols}

If you have any questions regarding the code, paper details, or other aspects of this work, you are very welcome to contact the authors at \email{guibinz@outlook.com} or via raising a \href{https://github.com/bingreeky/MemEvolve}{Github} issue.